\documentclass[lettersize,journal]{IEEEtran}
\usepackage{amsmath,amsfonts}

\usepackage{algorithmic}
\usepackage{array}
\usepackage[caption=false,font=normalsize,labelfont=sf,textfont=sf]{subfig}
\usepackage{textcomp}
\usepackage{stfloats}
\usepackage{url}
\usepackage{verbatim}
\usepackage{graphicx}
\def\BibTeX{{\rm B\kern-.05em{\sc i\kern-.025em b}\kern-.08em
    T\kern-.1667em\lower.7ex\hbox{E}\kern-.125emX}}
\usepackage{balance}

\usepackage{cite}
\usepackage{algorithm}
\usepackage{dsfont}

\usepackage{booktabs}

\usepackage{amssymb}
\usepackage[normalem]{ulem}
\usepackage{arydshln}

\newcommand{\best}[1]{\textbf{#1}}
\newcommand{\second}[1]{%
  \begingroup
  \renewcommand{\ULdepth}{2pt}%
  \uline{#1}%
  \endgroup
}

\begin{document}
\title{Label Granularity Skew in Federated Learning with Hierarchical Image Classification
}

\author{
    Jaeheon~Kim,
    Hokeun~Kim,~\IEEEmembership{Member,~IEEE},
    and Bong~Jun~Choi,~\IEEEmembership{Senior~Member,~IEEE} 
\thanks{Jaeheon Kim and Bong Jun Choi are with the School of Computer Science and Engineering,
Soongsil University, Seoul, Republic of Korea (e-mail: teeraiser@soongsil.ac.kr; davidchoi@soongsil.ac.kr).}
\thanks{Hokeun Kim is with the School of Computing and Augmented Intelligence,
Arizona State University, Tempe, United States (e-mail: hokeun@asu.edu).}
}

\markboth{}%
{Jaeheon Kim \MakeLowercase{\textit{(et al.)}: Label Granularity Skew in Federated Learning with Hierarchical Image Classification}}

\maketitle

\begin{abstract}
Federated learning enables privacy-preserving collaboration across distributed devices without centralizing local data. However, clients may differ not only in data distributions but also in domain knowledge and annotation capabilities. In this paper, we introduce label granularity skew, a new form of statistical heterogeneity in federated hierarchical classification, in which clients provide taxonomy-consistent labels at different levels of detail within a shared class hierarchy.
To model this heterogeneity, we generate client-specific local label hierarchies using a probabilistic relational neighbor classifier and construct a WordNet-guided hierarchy via silhouette score-based coarsening. Our analysis shows that strongly coupled hierarchical models are sensitive to incomplete supervision, while the conditional softmax classifier is more robust. Based on this insight, we propose Branch-wise Decoupled Fine-Tuning (BDFT) and its federated version, FedBDFT, which fine-tune branch-wise classifiers and aggregate them through federated optimization.
Experiments on CIFAR-100, TinyImageNet, and ImageNet show that FedBDFT substantially improves robustness under severe label granularity skew, with average gains of 27.9\% and 56.4\% at skewness levels of 0.6 and 0.9, respectively. Zero-shot results further indicate that FedBDFT better preserves hierarchical representations for unseen fine-grained classes. These findings demonstrate its effectiveness for federated hierarchical classification with heterogeneous label granularities.

\end{abstract}


\begingroup
\renewcommand\thefootnote{}
\footnotetext{
\hrule
\vspace{3pt}
This work has been submitted to the IEEE for possible publication. Copyright may be transferred without notice, after which this version may no longer be accessible.
}
\endgroup

\section{Introduction}

\IEEEPARstart{I}{n} traditional machine learning, the quality of supervised learning datasets is generally controlled by qualified experts. However, to ensure data diversity, some datasets are constructed by collecting data from diverse sources on a central server and annotating them through crowdsourcing~\cite{wah2011caltech, deng2009imagenet, inaturalist-2021}. In such cases, the quality of the labels provided by annotators may be imperfect due to differences in their expertise and interests. Campbell et al.~\cite{10.1093/biosci/biad051} analyzed the annotators of the iNaturalist dataset~\cite{inaturalist-2021} and found that 50\% were taxonomic specialists who focused on specific taxa, while 65\% were geographic specialists who labeled images from geographically proximate regions. Chang et al.~\cite{Chang_2021_CVPR} conducted a human study on the CUB-200-2011 dataset~\cite{wah2011caltech} and found that participants preferred different levels of label granularity depending on their level of bird expertise. Nevertheless, since such datasets are publicly accessible, these issues can be mitigated through cross-validation and collaborative consensus over labels. 

Federated learning (FL) enables distributed model training without centralizing data, thereby enhancing privacy by allowing participants to keep sensitive data on their local devices, but inherently amplifies heterogeneity across clients in terms of their data distributions, models, communication links, devices, and, crucially, label quality and consistency~\cite{ye2023heterogeneous, pei2024review}. Moreover, since clients' datasets are not publicly accessible, their labeling quality and consistency depend on the expertise and interests of individual clients. In particular, clients may be unable to provide labels at multiple levels of granularity because of differences in their knowledge or interests. In such cases, restricting training participation to clients with complete knowledge makes it difficult to obtain sufficiently diverse training data. In this direction, several prior studies have investigated hierarchical classification with partial labels \cite{10.1145/3583780.3614912, 9879829}. However, their formulations remain limited to centralized scenarios and instance-wise label incompleteness; thus, they do not adequately characterize the federated setting, in which client-specific label distributions may induce more structured hierarchical missingness, i.e., the absence of training samples for certain classes. 

Therefore, we investigate hierarchical classification with partial labels in federated learning and introduce \textbf{Label Granularity Skew}, a novel type of statistical heterogeneity specific to hierarchical classification. The objective of federated learning under label granularity skew is to train a global hierarchical classifier that captures knowledge across the entire class hierarchy by aggregating partial supervision from clients whose local datasets provide labels at different levels of granularity, thereby covering different portions of the hierarchy.

Meanwhile, prior work on hierarchical classification has largely focused on class hierarchies within narrow domains or on artificially simplified taxonomies, which do not adequately capture semantic relationships across diverse categories. Such limitations can introduce category confusion and hierarchical ambiguity, especially when label granularity differs across clients or annotation settings. To address this issue, we leverage WordNet to construct a more semantically grounded class hierarchy for hierarchical image classification. However, since the constructed WordNet hierarchy is often overly deep and imbalanced, we further propose a silhouette score-based hierarchy coarsening method that reduces label space complexity while preserving discriminative semantic structure.

\begin{figure*}[!t]
    \centering
    \includegraphics[width=1.0\textwidth]{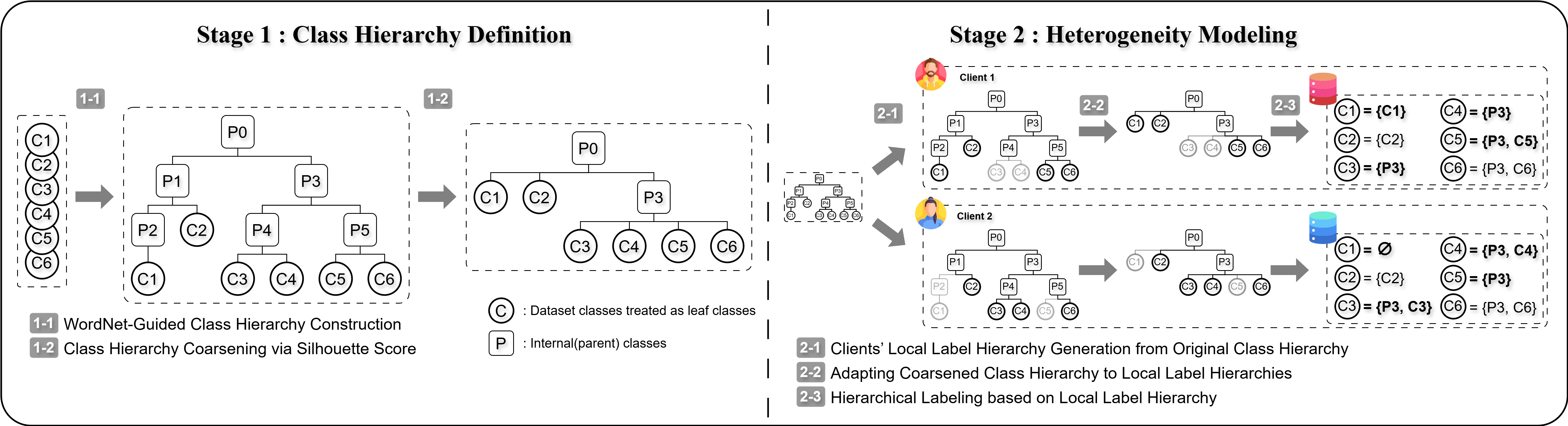}
    \caption{Overview of the environment setup for federated learning. The setup process consists of two steps: (Stage 1) class hierarchy definition and (Stage 2) client heterogeneity modeling. In the class hierarchy definition stage, dataset classes are matched to WordNet classes to construct the original class hierarchy ($\mathcal{H}_\text{full}$), which is then transformed into a coarsened class hierarchy ($\mathcal{H}$) using a coarsening algorithm (Algorithm~\ref{algo:class_hier_coarsening}). In the heterogeneity modeling stage, known and unknown classes are determined for each client based on $\mathcal{H}_\text{full}$ (Algorithm~\ref{algo:gen_local_label_hier}). The corresponding local label hierarchy is then constructed by removing the unknown classes from $\mathcal{H}$, and hierarchical labels are reassigned accordingly.}
    \label{fig:overview}
\end{figure*}

Furthermore, we observed that most existing hierarchical classification models are ineffective under label granularity skew. Flat-based classification models~\cite{NEURIPS2022_727855c3} cannot handle coarsely labeled data, while tightly coupled hierarchical classification models~\cite{brust2019integrating, NEURIPS2022_727855c3} tend to propagate noise and missing specificity throughout the taxonomy, hindering global convergence.
We experimentally verified that partial supervision of a conditional softmax classifier (\emph{cond softmax})~\cite{redmon2017yolo9000} is remarkably effective for handling label granularity skew. Motivated by this observation, we propose \textbf{Branch-wise Decoupled Fine-tuning (BDFT)}, which fine-tunes branch-wise local classifiers decoupled from a pretrained \emph{cond softmax} to further amplify this capability, and its federated algorithm FedBDFT.
We observe that state-of-the-art models designed to learn hierarchical features under strong structural constraints achieve high performance in less heterogeneous environments. However, in severe heterogeneous settings, such designs become increasingly vulnerable to the adverse effects of incomplete hierarchical labels. In contrast, \emph{cond softmax} and BDFT, owing to their weak inter-branch coupling and independent learning structure, achieve average performance gains of 27.9\% and 56.4\% over the baselines at skewness levels of 0.6 and 0.9, respectively.

This paper presents the following contributions:
\begin{itemize}
  \item We introduce \textbf{Label Granularity Skew}, a novel form of heterogeneity in federated hierarchical classification arising from differences in label granularity across clients with distinct areas of expertise. To model this heterogeneity, we propose a local label hierarchy generation algorithm based on the \textbf{Probabilistic Relational Neighbor (pRN)}~\cite{macskassy2003simple}.
  \item To define hierarchical relationships among classes more explicitly and precisely, we employ WordNet~\cite{miller1995wordnet}, a knowledge graph that reflects a realistic taxonomy. Furthermore, we propose \textbf{WordNet-guided hierarchy coarsening via silhouette score} to mitigate the expansion of the label space.
  \item We experimentally evaluated existing hierarchical classification models under the label granularity skew setting and found that the \textbf{conditional softmax classifier (\emph{cond softmax})}~\cite{redmon2017yolo9000} is the most effective. In addition, we introduce \textbf{Branch-wise Decoupled Fine-tuning (BDFT)} and \textbf{FedBDFT} to improve the effectiveness of \emph{cond softmax}. This method yields the best performance under label granularity skew.
\end{itemize}

This study introduces a new type of heterogeneity, termed label granularity skew, and proposes an approach that leverages WordNet, a well-established lexical knowledge graph, as a class hierarchy. Fig.~\ref{fig:overview} provides an overview of the procedure for defining the class hierarchy and modeling each client’s heterogeneity before the training stage.

The remaining part of this paper is organized as follows: Section~\ref{sec:problem_definition} defines the hierarchical classification problem and introduces the notation used throughout the paper. Section~\ref{sec:literature_reviews} reviews related work on federated learning under heterogeneous supervision, noisy labels, hierarchical image classification, and partial-label learning. Section~\ref{sec:dist_model} formalizes label granularity skew and describes the proposed procedure for generating client-specific local label hierarchies. Section~\ref{sec:exp-class_hierarchy} presents the WordNet-guided hierarchy construction and silhouette score-based coarsening method. Section~\ref{sec:lgs} introduces Branch-wise Decoupled Fine-Tuning and its federated extension, FedBDFT. Section~\ref{sec:exp} reports the experimental setup and evaluation results on benchmark image datasets. Finally, Section~\ref{sec:discussion} discusses the limitations and implications of the proposed framework, and Section~\ref{sec:conclusion} concludes the paper.

\section{Problem Definition}\label{sec:problem_definition}

\noindent Hierarchical classification is a type of multi-label classification based on a predefined class hierarchy. The class hierarchy can be represented as a \textbf{tree} or a \textbf{DAG}.
Under a tree-based class hierarchy $\mathcal{H}$, we define $Ch(e)$ as a set of child classes (subclasses) of class $e$; $Par(e)$ as a parent class (superclass) of $e$; $Sib(e)$ as the set of sibling classes of $e$; and $Anc(e)$ and $Desc(e)$ as the sets of ancestors and descendants of class $e$, respectively. $Path(e)$ is the unique sequence of classes on the path from the root to class $e$.

\subsection{Hierarchical Feature}\label{sssec:hier_feat}
\noindent Since parents and children are in an \emph{is-a} relationship within the class hierarchy, the features of a parent class are inherited by its child classes. In this nature, the set of features for class $e$, denoted $\mathcal{X}_{e}$, is represented as:
\begin{equation} \label{eq:common_feature}
\mathcal{X}_{e} = \bigcap_{c \in Ch(e)} \mathcal{X}_{c}.
\end{equation}

\begin{figure}[htbp]
    \centering
    \centerline{\includegraphics[width=0.45\textwidth]{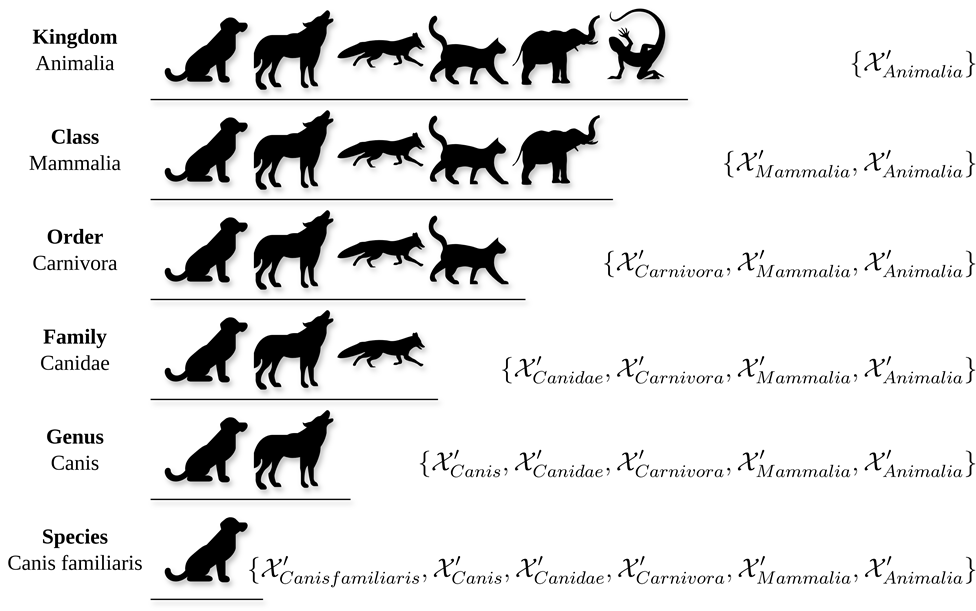}}
    \caption{An example of scientific classification and feature sets. The classes in the same row belong to the same category, and the expressions on the right indicate the set of visual features that they share.}
    \label{fig:scietific_classification}
\end{figure}

Then $\mathcal{X}_{e}$ is recursively given by:

\begin{equation} \label{eq:hierarchical_feature}
\begin{aligned}
\mathcal{X}_{e} & = (\mathcal{X}_{e} \setminus \; \mathcal{X}_{Par(e)}) \cup \mathcal{X}_{Par(e)} \\
& = (\mathcal{X}_{e} \setminus \; \mathcal{X}_{Par(e)}) \cup (\mathcal{X}_{Par(e)} \setminus \mathcal{X}_{Par(Par(e))}) \\
& \qquad\qquad\qquad\qquad\qquad\qquad\qquad \cup \; \mathcal{X}_{Par(Par(e))} \\
& = \cdots \\
& = \left(\bigcup_{\mathit{c} \in Anc(e)\setminus \{root\}} (\mathcal{X}_\mathit{e} \setminus \mathcal{X}_{Par(\mathit{e})})\right) \cup \mathcal{X}_{root}\\
& = \bigcup_{c \in Anc(e)} \mathcal{X}^\prime_\mathit{c}, \qquad \quad \because Par(root) = \emptyset, \mathcal{X}_\emptyset = \emptyset.
\end{aligned}
\end{equation}
where $\mathcal{X}^\prime_\mathit{e}$ denotes a \emph{distinctive feature set} for the class $e$, shared with $Desc(e)$ while distinguishing it from $Sib(e)$. And $\mathcal{X}^\prime_{root}$ is assumed $\emptyset$ since $root$ doesn't have siblings. Fig.~\ref{fig:scietific_classification} illustrates a partial scientific classification of the dog class. 

\subsection{Hierarchical Label}\label{sssec:hier_label}
\noindent To effectively train the hierarchical feature representation, labels that reflect the corresponding class hierarchy are required. Since $Path(e)$ is unique in a tree-structured class hierarchy, the hierarchical label $S_{e}$ can be defined as the set of labels of $c$ and its ancestors.

\begin{equation} \label{eq:common_label}
\begin{aligned}
S_{e} & = \{y_{e}\} \; \bigcup \; S_{Par(e)} \\
& = \cdots \\
& = \left( \bigcup_{\mathit{c} \in Anc(e)\setminus \{root\}} \{y_c\} \right) \cup S_{root}\\
& = \bigcup_{c \in Anc(e) \setminus \{root\}} \{y_c\}.
\end{aligned}
\end{equation}
Note that $S_{root}$ and $y_{root}$ are excluded because every class in the hierarchy is a descendant of the $root$.
Table~\ref{tab:symbols} lists the symbols and abbreviations used in this paper.

\begin{table*}[t]
\centering
\caption{List of Symbols and Abbreviations}
\label{tab:symbols}
\renewcommand{\arraystretch}{1.15}
\begin{tabular}{ll}
\toprule
\textbf{Symbol} & \textbf{Description} \\
\midrule
$\alpha$ & Label granularity skewness parameter. A larger $\alpha$ indicates more severe label granularity skew. \\
$\mathit{Anc}(e;\mathcal{H})$ & Set of ancestor classes of class $e$ in the class hierarchy $\mathcal{H}$, abbreviated as $\mathit{Anc}(e)$ \\
$\mathcal{C}(\mathcal{H})$ & Set of all classes in the class hierarchy $\mathcal{H}$, abbreviated as $\mathcal{C}$ \\
$\mathit{Ch}(e;\mathcal{H})$ & Set of child classes of class $e$ in the class hierarchy $\mathcal{H}$, abbreviated as $\mathit{Ch}(e)$ \\
$D$ & Dataset. \\
$D_e$ & Branch-specific dataset for the classifier associated with class $e$. \\
$D_e^{(k)}$ & Branch-specific dataset of client $k$ for the classifier associated with class $e$. \\
$\mathit{Desc}(e;\mathcal{H})$ & Set of descendant classes of class $e$ in the class hierarchy $\mathcal{H}$, abbreviated as $\mathit{Desc}(e)$ \\
$\mathcal{E}(\mathcal{H})$ & Set of intermediate classes mapped to local classifiers in BDFT, abbreviated as $\mathcal{E}$ \\
$\mathcal{H}$ & Coarsened class hierarchy used for hierarchical classification. \\
$\mathcal{H}_{\mathrm{full}}$ & Original WordNet-guided class hierarchy before coarsening. \\
$\mathcal{K}$ & Set of client participating in federated learning. \\
$\mathcal{L}(\mathcal{H})$ & Leaf class set in the class hierarchy $\mathcal{H}$, abbreviated as $\mathcal{L}$ \\
$\mathit{Lvs}(e;\mathcal{H})$ & Set of leaf descendant classes of class $e$ in the class hierarchy $\mathcal{H}$, abbreviated as $\mathit{Lvs}(e)$ \\
$\mathit{Par}(e;\mathcal{H})$ & A parent class of class $e$ in the class hierarchy $\mathcal{H}$, abbreviated as $\mathit{Par}(e)$ \\
$\mathit{Path}(e;\mathcal{H})$ & Unique sequence of classes from the root to class $e$ in the class hierarchy $\mathcal{H}$, abbreviated as $\mathit{Path}(e)$ \\
$S$ & Hierarchical label set. \\
$S_e$ & Hierarchical label set of class $e$, consisting of the labels of $e$ and its ancestors. \\
$S_e^{(k)}$ & Hierarchical label of class $e$ according to client $k$'s local label hierarchy. \\
$\mathit{Sib}(e)$ & Set of sibling classes of class $e$. \\
$W$ & Weighted adjacency matrix derived from inter-class Wu-Palmer similarity. \\
$X$ & Input set. \\
$X_e$ & Input set of class $e$. \\
$\mathcal{X}_e$ & Feature set of class $e$. \\
$\mathcal{X}'_e$ & A distinctive feature of class $e$, shared with its descendants and distinguished from siblings. \\
\bottomrule
\end{tabular}
\end{table*}

\section{Related Work}\label{sec:literature_reviews}

\subsection{Federated Learning under Heterogeneous Supervision}

\noindent A large body of federated learning research has studied heterogeneity across clients, including statistical, system, and model heterogeneity. In particular, recent work has considered supervision mismatch across clients, including differences in task definition, label availability, and label space composition. For example, mixed-label-space or mixed-type-label FL studies address the case where different institutions annotate data with different criteria or operate on partially overlapping label spaces \cite{11129977}. Similarly, task-heterogeneous FL has been explored in multi-label medical imaging, where clients may supervise only a subset of categories or tasks \cite{Sun_FedMLP_MICCAI2024}. These studies are closely related to our setting in that they relax the conventional assumption of globally consistent supervision.

However, our problem is not simply a mismatch of label spaces across clients. In our setting, clients share the same underlying semantic classes, but they may annotate the \emph{same} instance at different levels of a hierarchy according to their knowledge base. That is, the supervision discrepancy is induced by \emph{taxonomy-consistent variation in label granularity}, rather than by arbitrary task mismatch or institution-specific label schemas. This distinction is especially important in hierarchical image classification, where a coarse ancestor label is not incorrect, but incomplete relative to the finest-grained class. Existing FL studies on hierarchical image classification, such as FedTH, consider tree-structured prediction in the federated setting but do not explicitly model client-specific differences in hierarchical supervision granularity \cite{kim2022fedth}. Our work addresses this missing setting and formalizes it as \emph{label granularity skew}.

\subsection{Noisy and Imperfect Labels in Federated Learning}

\noindent Another relevant research direction is FL with noisy or imperfect labels. Prior studies have shown that label noise can significantly degrade convergence and generalization in federated optimization, especially when noise rates vary across clients \cite{10349830, 10.24963/ijcai.2023/492, 10447823}. These methods typically assume that observed labels are corrupted versions of latent ground-truth labels, and therefore focus on estimating clean labels, correcting label corruption, or improving robustness to client-wise noise heterogeneity.

Our setting differs fundamentally from this line of work. Under label granularity skew, the observed label is often a valid ancestor of the ground-truth class in the hierarchy, rather than a randomly corrupted or semantically incorrect label. Formally, the supervision can be interpreted as a taxonomy-consistent projection of the latent fine-grained label onto a client-specific local hierarchy. Therefore, the challenge is not merely label denoising, but learning under \emph{structured and semantically valid yet incomplete supervision}. This difference also explains why methods designed for conventional noisy-label FL are not directly sufficient: they treat discrepancies from the finest-grained label as errors, whereas in our setting, such discrepancies may reflect legitimate coarse-level knowledge.

\subsection{Hierarchical and Multi-Granularity Image Classification}

\noindent Hierarchical image classification has been extensively studied in the context of centralized learning. Early and recent methods incorporate class hierarchies through hierarchy-aware losses, parameter sharing, conditional prediction, or structured inference \cite{10.1007/978-3-030-41404-7_1, NEURIPS2022_727855c3, pmlr-v80-wehrmann18a, 9156542, 9151045}. These approaches exploit parent--child relations to improve semantic consistency and to evaluate predictions beyond exact-match accuracy.

Another closely related line of work studies multi-granularity recognition in computer vision. Several works recognize that different annotators or recognition settings may prefer coarse- or fine-grained categories depending on expertise \cite{Chang_2021_CVPR, 9879829}. This literature is highly relevant to our motivation, as it acknowledges that granularity itself is not fixed and may depend on the annotator's knowledge. 
Nevertheless, existing studies are largely centralized and assume that such granularity differences can be handled within a single training environment with globally accessible data, treating the resulting mismatch as stemming from randomness rather than structural noise.
They do not consider the federated case, where client data and client-specific label hierarchies remain local. 

Our work differs from previous studies on hierarchical classification in two ways. First, we examine a federated setting in which hierarchical supervision is inconsistent across clients. Second, we found that methods imposing strong structural coupling across the hierarchy are more vulnerable to severe granularity mismatch, whereas formulations with weaker coupling are substantially more robust. These observations motivate our branch-wise decoupled design, which led to the proposal of BDFT and FedBDFT.

\subsection{Partial-Label Learning and Hierarchy-Aware Partial Supervision}

\noindent Our work is also related to partial-label learning (PLL), which considers training examples associated with incomplete or ambiguous supervision \cite{10.5555/3524938.3525541, wang2022pico}. In standard PLL, an instance is typically given a candidate label set containing the true label, and the goal is to disambiguate that set during training. More recent studies have incorporated structural priors into PLL, including hierarchy-aware approaches that exploit taxonomic relations among labels \cite{10943569}. In addition, hierarchical multi-label classification with partial labels has been studied in centralized settings, including cases where the hierarchy itself may be partially unknown \cite{10.1145/3583780.3614912}.

Despite this connection, our problem is not equivalent to a conventional PLL. In PLL, ambiguity is usually defined at the instance level, and the candidate set is often treated as an uncertain superset of possible true labels. In contrast, our setting is induced by \emph{client-wise differences in local knowledge boundaries} over a shared hierarchy. The missing specificity is therefore structured by the hierarchy and systematically correlated with the client, not merely with individual instances. Moreover, in federated learning, client-specific label granularity can result in the effective absence of training samples for certain fine classes on some clients, creating an additional optimization difficulty that does not arise in centralized partial-label settings. Our formulation thus lies at the intersection of hierarchical learning, partial supervision, and federated heterogeneity, while remaining distinct from each of them.

\subsection{Position of This Work}

\noindent In summary, prior work has studied heterogeneous supervision in FL, noisy-label FL, hierarchical image classification, and partial-label learning. However, the specific setting where clients annotate the same semantic classes at different hierarchical levels based on their local expertise has received little attention. We address this gap by formalizing \emph{label granularity skew}, proposing a realistic client-side local hierarchy generation mechanism, and developing a federated training method that is robust to hierarchy-consistent but client-dependent labels incompleteness.

\section{Label Granularity Skew}\label{sec:dist_model}

\subsection{Definition of Label Granularity Skew}\label{ssec:lgs}
\noindent In centralized machine learning, the data, the corresponding labels, the model, and the training hardware are controlled by a central server. In contrast, federated learning distributes both the training participants and data ownership, leading to heterogeneity across various elements. Heterogeneity is commonly categorized as statistical, model, communication, and device heterogeneity; among these, statistical heterogeneity—arising from differences in data distributions—can be further divided into label skew, feature skew, quality skew, and quantity skew~\cite{ye2023heterogeneous}.
A subtype of label skew, \textbf{label preference skew}, refers to scenarios in which clients may assign different labels to the same data depending on their preference. In federated learning, unlike in centralized machine learning, each client's data is hidden from the central server; as a result, the server cannot observe their label distribution.

\begin{figure}[htbp]
    \centerline{\includegraphics[width=0.49\textwidth]{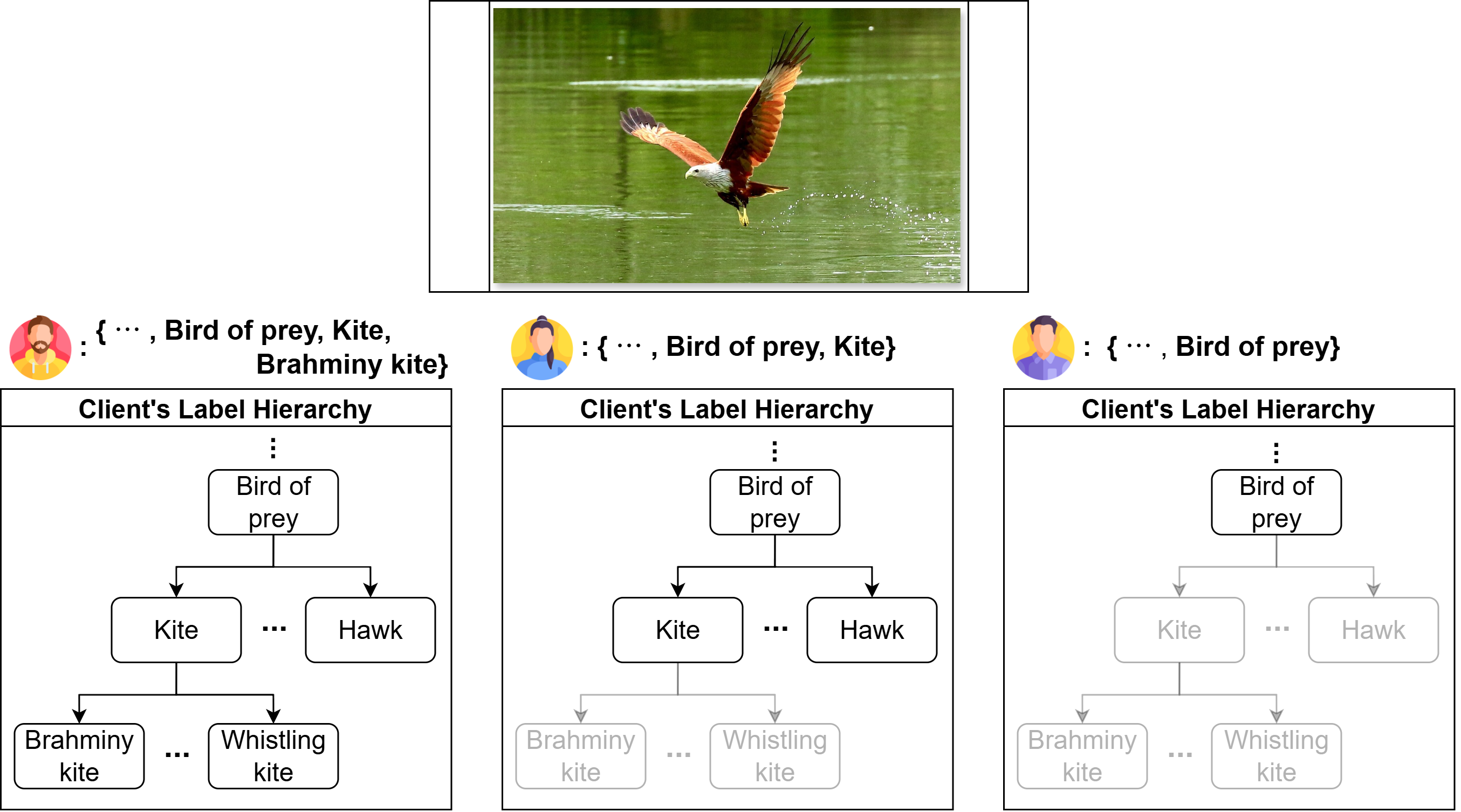}}
    \caption{Label Granularity Depends on Clients' Knowledge. The gray-shaded classes indicate the classes that are unknown to each client.}
    \label{fig:client_expertise}
\end{figure}

In hierarchical classification, even though a client is unfamiliar with a class’s subcategories, they can still provide coarse-grained labels, because a hierarchical label comprises a set of labels at multiple granularities along the class hierarchy for a single data point. However, this incompleteness leads to inconsistency between the ground truth and the labels of classes.
For example, for the class $\textit{Brahminy kite}$, if a client $k$ has sufficient knowledge up to $\textit{Brahminy kite}$, the hierarchical label is $S^{(k)}_{\textit{Brahminy kite}} = \{y_{_\textit{Animal}}, y_{_\textit{Chordate}}, y_{_\textit{Bird}}, y_{_\textit{Bird of prey}}, y_{_\textit{Kite}}, y_{_\textit{Brahminy kite}}\}$; if the client’s knowledge extends only to $\textit{Bird}$, then $S^{(k)}_{\textit{Brahminy kite}} = S^{(k)}_{c \in Desc(\textit{Bird})\cup\{\textit{Bird}\}} = \{y_{_\textit{Animal}}, y_{_\textit{Chordate}}, y_{_\textit{Bird}}\}$. Fig.~\ref{fig:client_expertise} illustrates how label granularity varies with clients’ knowledge represented in the label hierarchy.

\textbf{Label hierarchy} denotes the hierarchy used for hierarchical labeling, reflecting the knowledge each entity has, and is represented as a subtree of the class hierarchy defined for the classification task. The label hierarchies associated with clients are termed \textbf{local label hierarchies}, while the server-side label hierarchy, identical to the original class hierarchy, is termed the \textbf{global label hierarchy}.

Therefore, we define \textbf{label granularity skew}, a type of label preference skew in federated hierarchical classification in which clients, depending on their \emph{local label hierarchy}, exhibit varying label granularity in the hierarchical labels they provide. 
The distributions of the feature and the finest-grained label are equivalent to flat classification label preference skew: $P_i (x) = P_j (x)$, but $P_i (y \vert x) \neq P_j (y \vert x)$ for clients $i$ and $j$.

However, with the hierarchical label structure $S$, there exists a class $e$ such that
\begin{equation}
\begin{aligned}
P_i(X_e) &= P_j(X_e), \; P_i(S^{(i)}_e \mid X_e) \neq P_j(S^{(j)}_e \mid X_e), \\
&\text{whereas } P_i(S^{(i)}_{Par(e)} \mid X_e) = P_j(S^{(j)}_{Par(e)} \mid X_e),
\end{aligned}
\end{equation}
where
\begin{equation}
X_e = \{\, x \in X \mid (x,S) \in D,\, S_e \subseteq S\,\}.
\end{equation}

\subsection{Modeling of Local Label Hierarchy}
\noindent In the \textbf{label granularity skew} scenario, the set of known classes within the class hierarchy varies across clients based on their expertise. By the nature of the hierarchical structure, if class $c$ is known, then all ancestor classes $Anc(c)$ are also known; conversely, if class $c$ is unknown, then all descendant classes $Desc(c)$ are also unknown.
A straightforward method for constructing local label hierarchies is to randomly designate a subset of leaf classes as unknown. However, this method fails to realistically capture practical settings, as it confines unknown classes to the leaf level while implicitly assuming that all internal classes remain known.
Another possible strategy is to randomly assign a subset of leaf classes as known. Yet, owing to the hierarchical nature, any internal class whose descendant leaf classes are unknown must also be considered unknown. This assumption is similarly unrealistic.
For instance, even if a client lacks the knowledge necessary to differentiate among fine-grained dog breeds, it may still be able to recognize the higher-level class dog, provided that it possesses sufficient knowledge of semantically adjacent or hierarchically related classes.

Thus, we adapt the \textbf{probabilistic relational neighbor classifier (pRN)}~\cite{macskassy2003simple} to construct a realistic local label hierarchy for \textbf{label granularity skew}.
At initialization, we assign a proportion $\alpha$ of the leaf classes as \emph{unknown}, and the remaining $1 - \alpha$ and their ancestors as \emph{known}. All unassigned classes are marked as \emph{unlabeled}. The class probabilities are initialized as:
\begin{equation}
\label{eq:lgs_prob}
p = \{p_0, p_1, \cdots, p_{|\mathcal{K}|}\}, \text{where} \; p_i = 
\begin{cases}
0       & \text{if } i \in \textit{unknown},\\
1       & \text{if } i \in \textit{known},\\
0.5     & \text{otherwise.}
\end{cases}
\end{equation}
where $\mathcal{K}$ denotes the class set in the original class hierarchy $\mathcal{H}_\text{full}$.

Next, the \emph{unlabeled} nodes are randomly permuted, and the probability that is \emph{Known} of each node  is sequentially updated using:
\begin{equation}
\label{eq:prn}
P(Y_i=1) = \frac{1}{\sum_{(i, j) \in E}{W(i, j)}}{\sum_{(i, j) \in E}{W(i, j) \cdot P(Y_j=1),}}
\end{equation}
where $i$ and $j$ denote nodes, and $W(i, j)$ is a weighted adjacency matrix, which in our model we define using the Wu-Palmer distance.
This iteration is repeated until the node label probabilities converge.

\begin{figure}[htbp]
  \centering
  \subfloat[]{\includegraphics[width=0.49\linewidth]{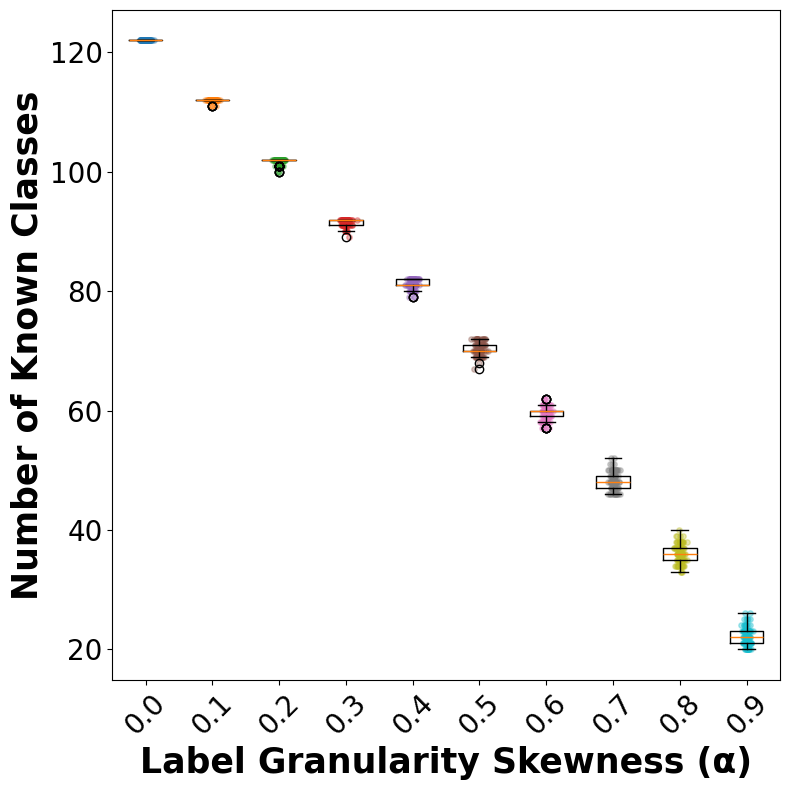}}\hfill
  \subfloat[]{\includegraphics[width=0.49\linewidth]{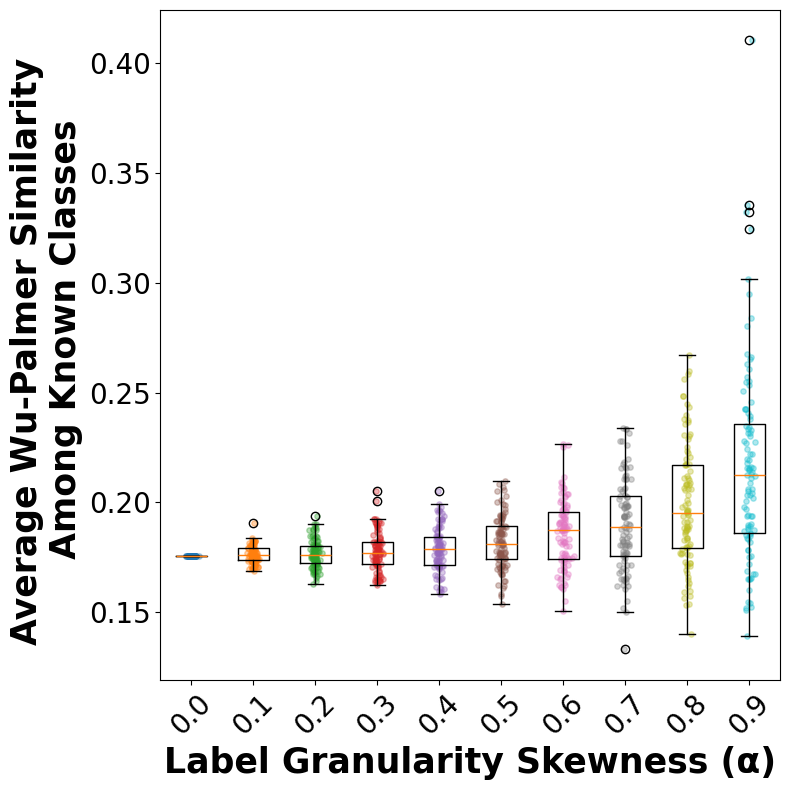}}\hfill
  \caption{Properties of clients' local label hierarchies at different levels of label granularity skewness on the CIFAR-100 dataset.
(a) Number of known classes in each local label hierarchy. (b) Average Wu-Palmer similarity among known classes within each local label hierarchy.}
  \label{fig:hierarchy_attributes}
\end{figure}

\begin{figure}[htbp]
  \centering
  \subfloat[]{\includegraphics[width=0.33\linewidth]{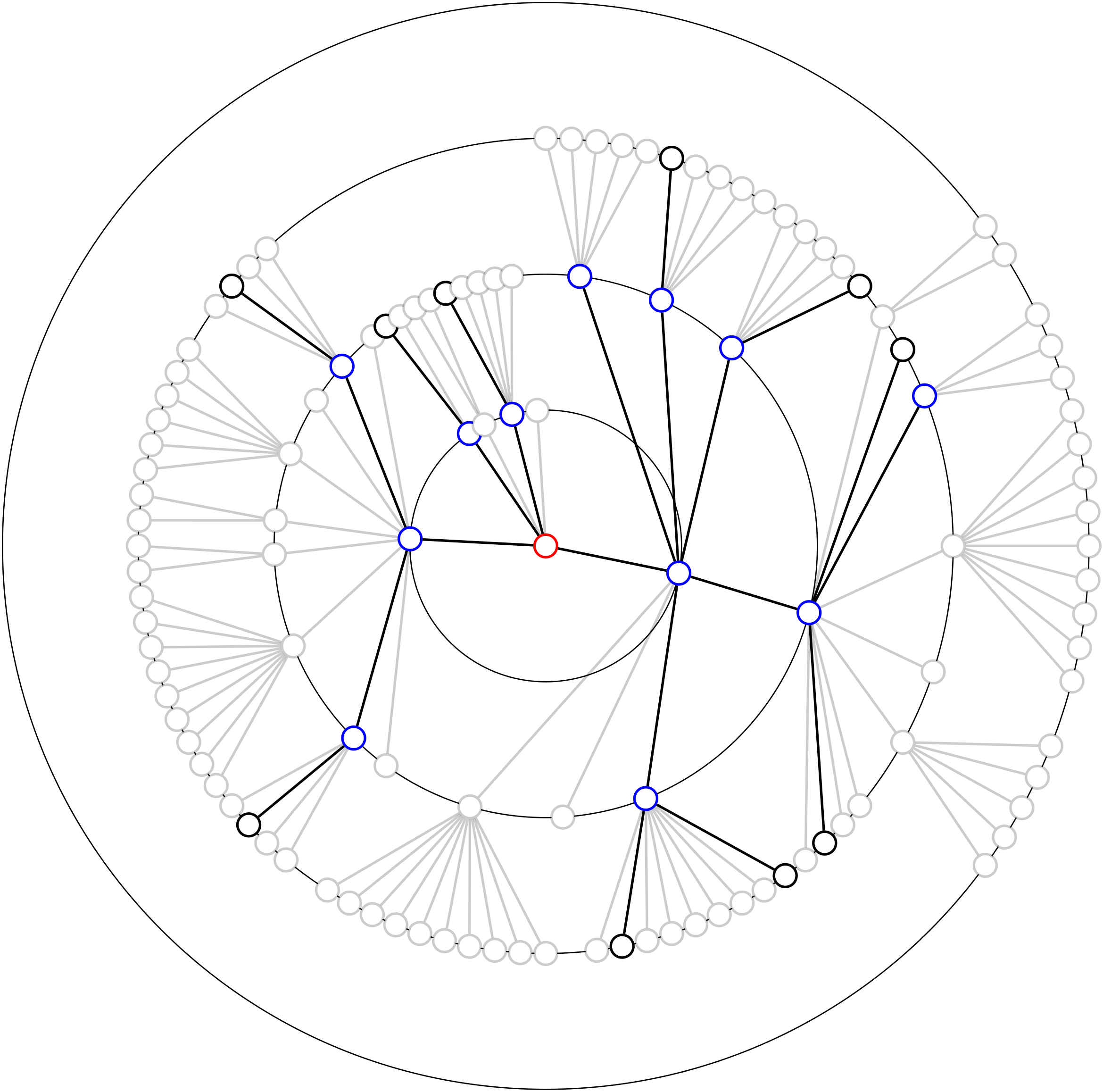}}\hfill
  \subfloat[]{\includegraphics[width=0.33\linewidth]{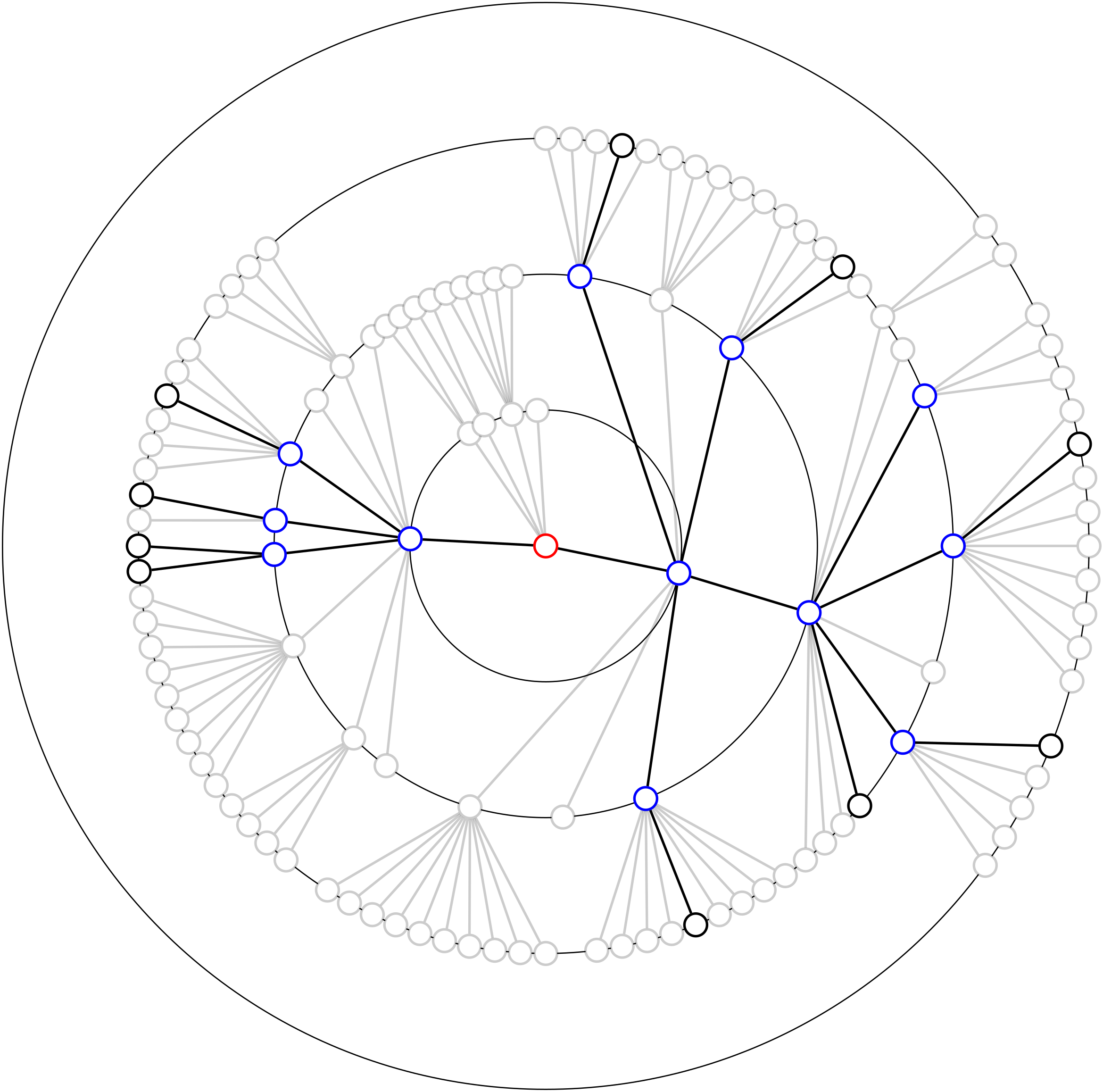}}\hfill
  \subfloat[]{\includegraphics[width=0.33\linewidth]{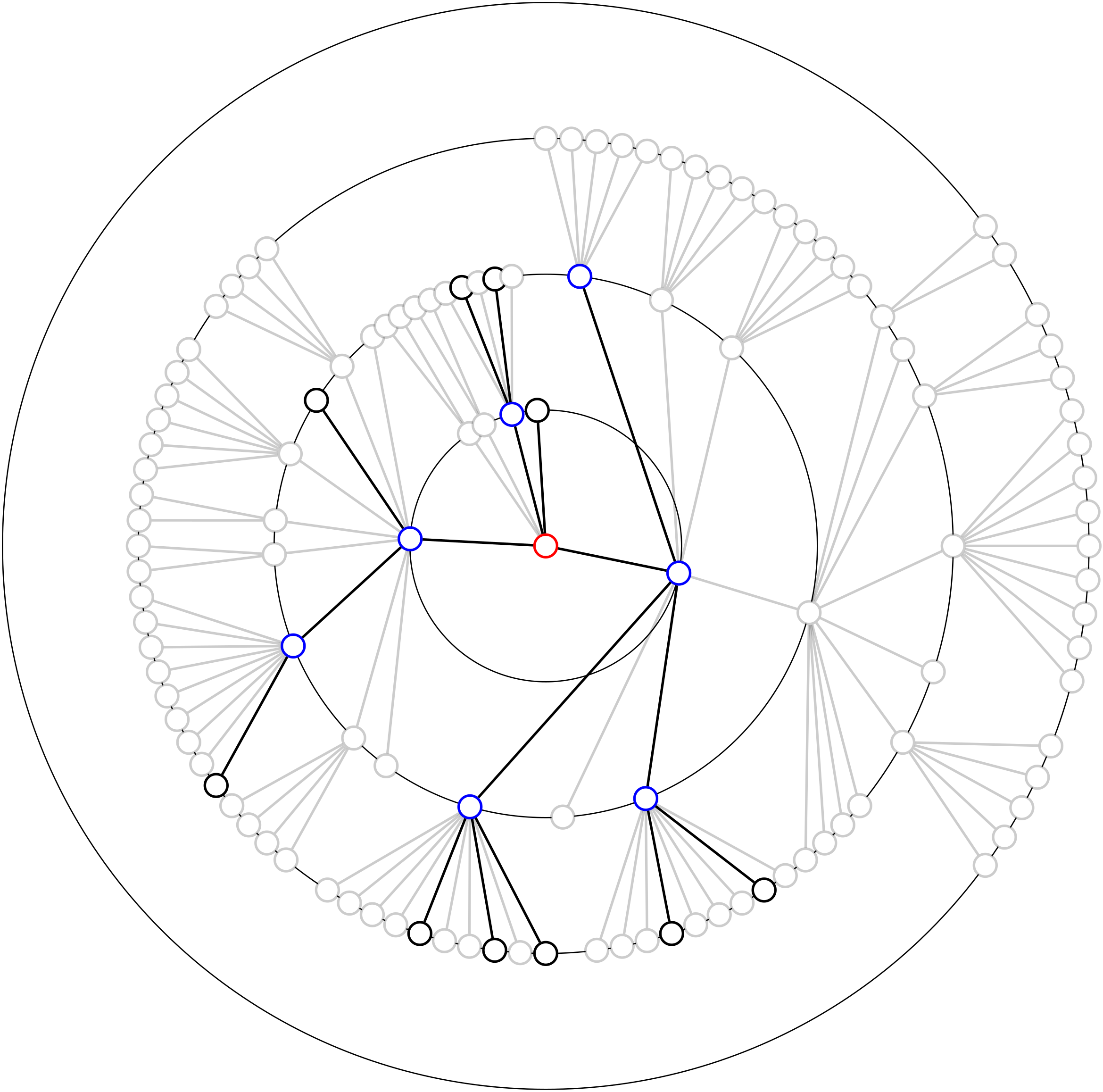}}\hfill

  \caption{(a–c) show examples of clients' local label hierarchy on the CIFAR-100 dataset with $\alpha=0.9$, visualized with radial trees. Gray nodes denote unknown classes, while colored nodes denote known classes. The red node is the root class, blue nodes are intermediate classes, and black nodes are leaf classes. Concentric circles around the red node indicate class depths in the coarsened class hierarchy.}
  \label{fig:knowledge_model_example}
\end{figure}

Algorithm~\ref{algo:gen_local_label_hier} presents the procedure for generating clients' local label hierarchy under the label granularity skew scenario. Fig.~\ref{fig:hierarchy_attributes} shows the properties of generated local label hierarchies at different levels of label granularity skewness ($\alpha$) on the CIFAR-100 dataset, and Fig.~\ref{fig:knowledge_model_example} visualizes representative hierarchies for $\alpha=0.9$.

\begin{algorithm}[!t]
\caption{Generating Local Label Hierarchies}
\label{algo:gen_local_label_hier}
\begin{algorithmic}[1]

\REQUIRE Class set $\mathcal{C}$; client set $\mathcal{K}$; parameter $\alpha \in [0,1-\frac{1}{\vert \mathcal{C} \vert }]$; leaf class set $\mathcal{L}$; weighted adjacency matrix $W$
\STATE Evenly partition $\mathcal{L}$ into $|\mathcal{C}|$ disjoint subsets $\{\mathcal{L}^{(1)}, \mathcal{L}^{(2)}, \ldots, \mathcal{L}^{(|\mathcal{K}|)}\}$
\FOR{$c \gets 1$ \textbf{to} $|\mathcal{K}|$}
    \item[] \hspace{\algorithmicindent}\text{// Initialize labels for labeled nodes and probabilities}
    \item[] \hspace{\algorithmicindent}\text{   for unlabeled nodes according to skewness ($\alpha$)}
    \STATE $s \gets \lceil (1-\alpha)\cdot |\mathcal{L}| \rceil - |\mathcal{L}^{(k)}|$
    \IF{$s > 0$}
        \STATE $S \gets \textsc{SimpleRandomSampling}\big(\mathcal{L} \setminus \mathcal{L}^{(k)},\, s\big)$
        \STATE $\mathcal{L}^{(k)} \gets \mathcal{L}^{(k)} \cup S$
    \ENDIF
    \FOR{$e \in \mathcal{L}^{(k)}$}
        \STATE $Known \gets Known \cup Path(e)$
    \ENDFOR
    \STATE ${Unknown} \gets \mathcal{L} \setminus \mathcal{L}^{(k)}$
    \STATE ${Unlabeled} \gets \mathcal{K} \setminus (Known \cup Unknown)$
    \STATE Initiate $\mathbf p$ with Eq.~(\ref{eq:lgs_prob}) \\ 
    \item[] \hspace{\algorithmicindent}\text{// perform node classification using pRN:}
    \REPEAT
        \STATE ${Unlabeled} \gets \textsc{Shuffle}({Unlabeled})$
        \FORALL{$i \in {Unlabeled}$}
            \STATE $\mathbf p_i \gets Eq.~(\ref{eq:prn})\big(i,\; \mathbf p,\; W\big)$
        \ENDFOR
    \UNTIL{$\mathbf{p} \text{ converges}$}
    \STATE $\mathbf Y^{(k)} = \mathds{1}\{\mathbf p \ge 0.5\}$
\ENDFOR

\RETURN $\{Y^{(1)}, Y^{(2)}, \cdots , Y^{(|\mathcal{C}|)}\}$
\end{algorithmic}
\end{algorithm}

\section{WordNet-Guided Hierarchy Coarsening via Silhouette Score}\label{sec:exp-class_hierarchy}

\noindent Existing studies have presented hierarchical models for classification that assume shallow label hierarchies with only two or three levels. These tasks generally consider hierarchical relations within a narrow class domain (e.g., Stanford Cars~\cite{krause2013collecting} contains only car categories, whereas CUB-200-2011~\cite{wah2011caltech} contains only bird categories) or abstract taxonomic structures introduced for experimental convenience (e.g., \emph{mushroom} is categorized as a subclass of \emph{fruit and vegetables} in CIFAR-100~\cite{krizhevsky2009learning}). 
Thus, such simplified tasks do not sufficiently capture semantic relationships among classes across diverse categories and may lead to category confusion and hierarchical ambiguity in the annotation process. To address these limitations, we adapt WordNet to model inter-class relationships more precisely, clarify category semantics, and further account for client-specific differences in label granularity.

However, the original WordNet-guided class hierarchy is often highly imbalanced, with most superclasses containing fewer than five subclasses and the hierarchy itself being excessively deep. Consequently, a hierarchical classification task based on the original WordNet-guided hierarchy would result not only in an excessively large label space but also in the model learning unnecessarily detailed hierarchical features.
Thus, we propose \textbf{WordNet-Guided Hierarchy Coarsening via Silhouette Score} to mitigate label space inflation while utilizing a discriminative WordNet-guided class hierarchy.
First, the original class hierarchy, $\mathcal{H}_{\text{full}}$, is constructed by traversing WordNet from the leaf classes and selecting the candidate path that exhibits the strongest semantic-visual alignment for hierarchical image classification. Next, to generate a coarsened class hierarchy $\mathcal{H}$, negligible classes are collapsed, and their parents are reassigned via graph coarsening. Negligibility is quantified using the \emph{Silhouette score}~\cite{rousseeuw1987silhouettes} of clusters with Wu-Palmer similarity~\cite{wu1994verb}.
Table~\ref{tb:expsclass_hier_info_imagenet} provides properties of the original class hierarchy $\mathcal{H}_{\text{full}}$ and the coarsened class hierarchy $\mathcal{H}$ in ImageNet, and Algorithm~\ref{algo:class_hier_coarsening} describes the class hierarchy coarsening algorithm.

\begin{table}[htbp]

\centering
\caption{Class Hierarchies' Information on ImageNet}
\label{tb:expsclass_hier_info_imagenet}
\renewcommand{\arraystretch}{1.5}
\begin{tabular}{|>{\centering}p{4cm}|c|c|}
\noalign{\smallskip}\noalign{\smallskip}

\hline
                                       & $\mathcal{H}_{\text{full}}$ & $\mathcal{H}$ \\
\hline
Total number of classes                & 1372             & 1066   \\
\hline
The number of internal classes                  & 372             & 66   \\
\hline
Maximum depth                          & 15              & 6    \\    
\hline
Average number of children of intermediate classes  & 3.69            & 15.91  \\
\hline
\end{tabular}
\end{table}

\begin{algorithm}[t]
\caption{WordNet-Guided Hierarchy Coarsening via Silhouette Score}
\label{algo:class_hier_coarsening}
\begin{algorithmic}[1]
\REQUIRE Original class hierarchy $\mathcal{H}_{\text{full}}$; weighted adjacency matrix $W$; maximum number of children $M$

\STATE $\mathcal{H} \gets \mathcal{H}_{\text{full}}$
\FOR{\textbf{each} class $u$ \textbf{in} \textsc{PostOrderInternalClass}$(\mathcal{H})$}
  \STATE ${Child}_{\text{orig}} \gets Ch(u;\mathcal{H})$ \quad {// get children set of $u$ in $\mathcal{H}$}
  \STATE $I \gets \{ v \in {Child}_{\text{orig}} \mid v \text{ is an internal class in } \mathcal{H}\}$
  \IF{$I = \emptyset$} \STATE \textbf{continue} \ENDIF
  \STATE $\text{bestScore} \gets -1$; \ ${Child}_{\text{best}} \gets {Child}_{\text{orig}}$
  \STATE $\mathcal{L}^\prime \gets Lvs(u;\mathcal{H})$ {// get leaf descendants set of $u$ in $\mathcal{H}$}
  \STATE $W^\prime \gets W[\mathcal{L}^\prime \times \mathcal{L}^\prime]$
  \STATE $P \gets PowerSet(I)$
  \FOR{\textbf{each} subset $S \text{ in } P$}
     \STATE ${Child}^\prime \gets {Child}_{\text{orig}}$
     \FOR{\textbf{each} $v \in S$}
        \STATE ${Child}^\prime \gets \big({Child}^\prime \setminus \{v\}\big)\ \cup\ Ch(v;\mathcal{H})$
     \ENDFOR
     \IF{$|{Child}^\prime| > M$} \STATE \textbf{continue} \ENDIF
     \STATE ${labels} \gets \textsc{Group} \; \mathcal{L^\prime} \; \textsc{by common ancestor in} \; \mathcal{C^\prime}$
     \STATE $s \gets \textsc{SilhouetteScore}(W^\prime, {labels})$
     \IF{$s > \textit{bestScore}$}
        \STATE $\textit{bestScore} \gets s$; \ ${Child}_{\text{best}} \gets {Child}^\prime$
     \ENDIF
  \ENDFOR
  \STATE $Ch(u;\mathcal{H}) \gets {Child}_{\text{best}}$  {// set children set of $u$ in $\mathcal{H}$ to ${Child}_{\text{best}}$}
\ENDFOR
\RETURN $\mathcal{H}$
\end{algorithmic}
\end{algorithm}

\section{Branch-wise Decoupled Fine-Tuning in Federated Learning}\label{sec:lgs}
\noindent Label granularity skew is heterogeneity in federated hierarchical classification, arising when clients possess different label hierarchies, such that the same instance may be annotated at different levels of label resolution. Consequently, the training label may not coincide with the ground-truth class label. In particular, in the multi-label classification setting, negative labels may contain a mixture of \emph{true negative labels} and \emph{unobserved positive labels (i.e., false negative labels)} \cite{10.1145/3583780.3614912}. Existing hierarchical models learn hierarchical features under the assumption that all negative labels are true negatives. However, under label granularity skew, this assumption is violated, leading these models to learn from incomplete labels and potentially encode incorrect supervisory signals. Furthermore, class-wise partial labeling impedes local client models from correcting such errors.

We empirically show that the conditional softmax classifier (\emph{cond softmax})~\cite{redmon2017yolo9000} that captures hierarchical relationships with a more weakly coupled hierarchical loss is effective under label granularity skew in Section~\ref{ssec:exp-result}, and motivated by this finding, we propose \textbf{Branch-wise Decoupled Fine-Tuning (BDFT)}, decoupling a pretrained \emph{cond softmax} at every branch and further fine-tuning each branch separately, along with its federated learning algorithm, \textbf{FedBDFT}.

\subsection{Branch-wise Decoupled Fine-Tuning}\label{sec:arch_of_hmet}
\noindent The \emph{cond softmax} learns the decision boundaries between the sibling classes in each branch. It is composed of a shared feature extractor and branch-specific classification heads, and it jointly learns to capture hierarchical features over the entire hierarchy in a single end-to-end training process. Meanwhile, BDFT shifts the single hierarchical classification task originally handled by \emph{cond softmax} to branch-wise subdivided flat classification tasks, following the LCPN structure in which local classifiers are deployed at parent nodes.

\begin{figure}[htbp]
    \centering    \includegraphics[width=0.49\textwidth]{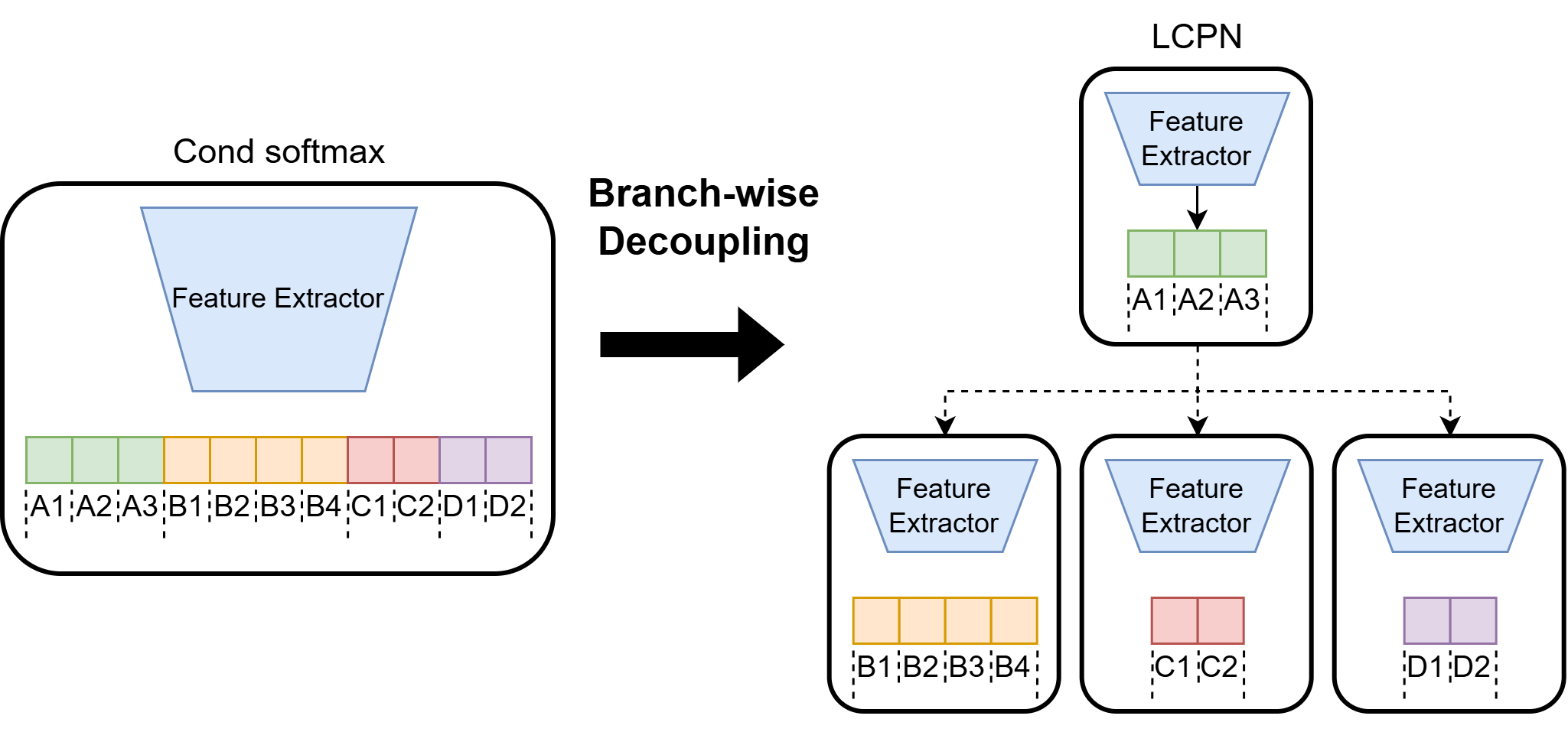}
    \caption{Branch-wise decoupling. Under a given class hierarchy, classes A1 to A3 are superclasses, and other class groups are subclasses of each superclass. The branch-specific classification heads of \emph{cond softmax} can be implemented using a linear layer, whereas the local classifiers in LCPN are architecturally decoupled.}
    \label{fig:exp-pt}
\end{figure}

Given the branch-wise equivalence between \emph{cond softmax} and LCPN, \emph{cond softmax} can be decoupled from the shared feature extractor and branch-specific classification heads to independent local classifiers composed of a branch-specific feature extractor and classification head. Fig.~\ref{fig:exp-pt} illustrates the branch-wise \emph{cond softmax} decoupling.

Consequently, the decoupled local classifiers address the branch-wise flat classification problem. Subsequently, each local classifier is individually fine-tuned to specialize in its respective task. To achieve this, the training dataset must be restructured from a multi-label format into a branch-specific single-label format. As stated in (\ref{eq:hierarchical_feature}), the feature representation of class $e$ corresponds to the intersection of the feature representations of the classes in $Desc (e)$. Therefore, $D_e$ can be restructured as follows:
\begin{equation} \label{eq:dataset_reconsistution}
\begin{split}
D_e &= \bigcup_{c \in Ch(e)} \{\, (x, y_c) \mid (x, S) \in D, y_c \in S\}. \\
\end{split}
\end{equation}
For example, images of both \textit{Retriever} and \textit{Shepherd} can be used to train a representation of \textit{Canidae} by labeling them as \textit{Canidae}. In this context, subproblems assigned at higher levels of the hierarchy capture general features, while others capture specific features. Along with the independence of each local classifier, this characteristic enables the model to effectively capture general features while retaining relevant information.

\subsection{Goal of FedBDFT}\label{sec:goal_of_hmet}
\subsubsection{Objective Function of BDFT}\label{ssec:erm_of_hmet}
\noindent Since each local classifier is an independent neural network, the empirical risk of a local classifier of class $\mathit{e}$ is defined as:
\begin{equation} \label{eq:expert_empirical_risk}
\begin{split}
\mathcal{L}_\mathit{e}(\theta_\mathit{e}) = \frac{1}{\vert D_{\mathit{e}}\vert} \sum_{(x, y) \in D_{\mathit{e}}} \ell_\mathit{e}(f_\mathit{e}(x, \theta_\mathit{e}), y).
\end{split}
\end{equation} 

Furthermore, since the local classifiers in the decoupled model are independent of one another and each is assigned a distinct subtask, the empirical risk of BDFT, which should be minimized, can be expressed as the sum of their empirical risks, defined as:
\begin{equation} \label{eq:architecture_empirical_risk}
\begin{split}
\mathcal{L}(\theta) & =  \sum_{\mathit{e}\in \mathcal{E}} \mathcal{L}_\mathit{e}(\theta_\mathit{e}) \\
& = \sum_{\mathit{e}\in \mathcal{E}}  \frac{1}{\vert D_{\mathit{e}}\vert} \sum_{(x, y) \in D_{\mathit{e}}} \ell_\mathit{e}(f_\mathit{e}(x, \theta_\mathit{e}), y)
\end{split}
\end{equation}
where $\mathcal{E}$ refers to a set of intermediate classes that are mapped to local classifiers in BDFT.

\subsubsection{Objective Function of FedBDFT} \label{ssec:in_federated_learning}
\noindent In contrast to conventional federated learning settings, which typically employ a monolithic and coupled single model, the decoupled model obtained through branch-wise \emph{cond softmax} decoupling comprises multiple independent neural networks, referred to as local classifiers. Minimizing the empirical risk of BDFT is equivalent to independently minimizing the empirical risk of each local classifier. Similarly, minimizing the empirical risk of FedBDFT corresponds to independently minimizing the empirical risk of all global local classifiers, as described in (\ref{eq:architecture_empirical_risk}). Notably, any federated algorithm can be adopted to aggregate the local classifiers in this context. In this study, we chose FedAvg~\cite{mcmahan2017communication} to focus on the effectiveness of BDFT. Thus, the empirical risk of each global local classifier in the federated setting can be described by
\begin{equation} \label{eq:expert_empirical_risk_in_federated_learning}
\begin{split}
\mathcal{L}_\mathit{e}(\theta_\mathit{e}) 
 &= \sum_{k \in \mathcal{K}} \frac{\vert D^{(k)}_{\mathit{e}}\vert}{\vert D_\mathit{e}\vert} \mathcal{L}_{\mathit{e}}(\theta^{(k)}_{\mathit{e}}).
\end{split}
\end{equation}

Finally, the objective of empirical risk minimization in FedBDFT can be defined as follows:

\begin{equation} \label{eq:fedtahme_objective_loss}
\begin{split}
\min_{\theta_{1:\vert\mathcal{E}\vert}} \sum_{\mathit{e} \in \mathcal{E}} \mathcal{L}_\mathit{e}(\theta_\mathit{e})
 &= \min_{\theta_{1:\vert\mathcal{E}\vert}} \sum_{\mathit{e} \in \mathcal{E}} \sum_{k \in \mathcal{K}} \frac{\vert D^{(k)}_{\mathit{e}}\vert}{\vert D_\mathit{e}\vert} \mathcal{L}_{\mathit{e}}(\theta^{(k)}_{\mathit{e}}).
\end{split}
\end{equation}

\subsection{Training Process of FedBDFT} \label{sec:train_in_fedhmet}
\begin{figure}[htbp]
    \centering    \includegraphics[width=0.45\textwidth]{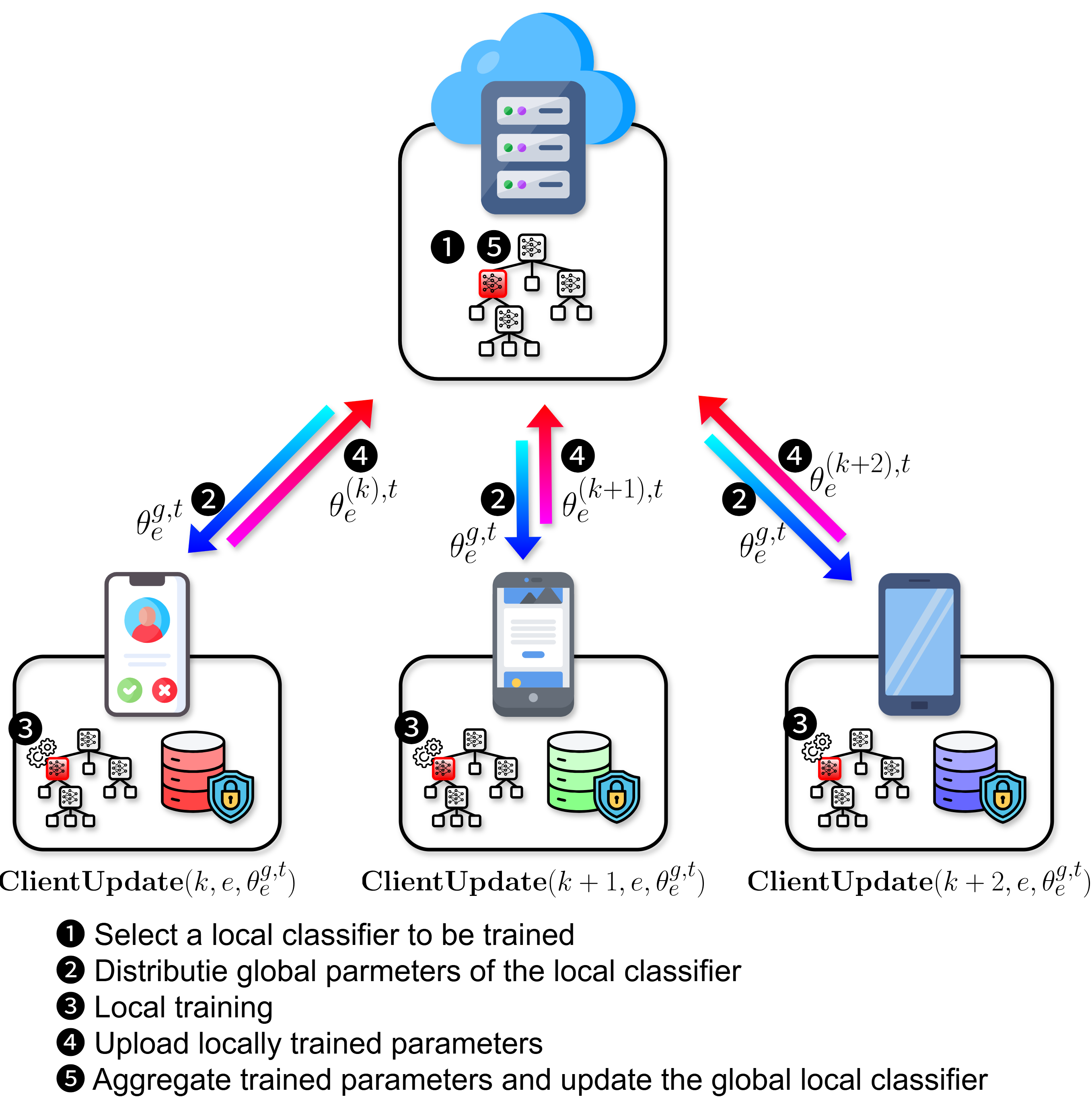}
    \caption{The overall training process of FedBDFT}
    \label{fig:training_fedhmet}
\end{figure}

\noindent Under this objective function, the training procedure does not simultaneously optimize the multiple independent local classifiers; instead, the server selects a single local classifier per round and trains it sequentially using a FedAvg-based approach. Specifically, the server distributes the global parameters of the classifier currently being trained to the participating clients, and each client performs local updates only for that classifier. The server then updates the global parameters of that classifier by taking a weighted average of the collected local parameters. By repeating this process for all local classifiers, the entire model is eventually trained sequentially. This decoupled optimization approach has the advantage of mitigating interference between classifiers while fully leveraging the aggregation mechanism of the standard FedAvg.
Fig.~\ref{fig:training_fedhmet} illustrates the overall FedBDFT training process, and its algorithm is presented in Algorithm~\ref{algo:training}.

\begin{algorithm}[htp]
\caption{Training Process of FedBDFT}
\label{algo:training}
\begin{algorithmic}[1] %

\REQUIRE The number of training rounds $T$; the number of local epochs $M$; the minimum number of training data $\gamma$; learning rate $\mu$; parameters of a global local classifier and client $k$'s local classifier $\mathit{e}$ at $t$-th round $\theta^{g, t}_{\mathit{e}}$, $\theta^{(k), t}_{\mathit{e}}$
\STATE \textbf{Server executes:}
    \FOR{$e \in \mathcal{E}$}
        \STATE Initialize $\theta^{g, 0}_{e}$
    \ENDFOR
    
    \FOR{each round $t$ = 0, 1, \ldots, $T-1$}
        \STATE $e \gets$ Select a local classifier to be trained in $\mathcal{E}$
        \STATE $S \gets$ Randomly select $K$ clients who $|D^{(k)}_{\mathit{e}}| \geq \gamma$
        \FOR{each client $k \in S$ \textbf{in parallel}}
            \STATE $\theta^{(k), t}_{\mathit{e}} \gets$ ClientUpdate$(k, \mathit{e}, \theta^{g, t}_{\mathit{e}})$
        \ENDFOR
        \STATE $\theta^{g, t+1}_{\mathit{e}} \gets \underset{{k \in \mathcal{S}}}{\sum} \frac{\vert D^{(k)}_{\mathit{e}}\vert}{\vert D_\mathit{e}\vert} \theta^{(k), t}_{\mathit{e}}$
    \ENDFOR
    
\STATE \textbf{ClientUpdate}($k, \mathit{e}, \theta_\mathit{e}$):
    \STATE $D^{(k)}_{\mathit{e}} \gets $ Reconstitute $(X_\mathit{e}, Y_\mathit{e})$ by Eq.~(\ref{eq:dataset_reconsistution})
    \FOR{each local epoch $m$ from 1 to $M$}
        \FOR{batch $(X_e, Y_e) \in D^{(k)}_{\mathit{e}}$}
            \STATE $\theta_\mathit{e} \gets \theta_\mathit{e} - \mu_\mathit{e} \nabla \ell_\mathit{e}(f_\mathit{e}(X_e, \theta_\mathit{e}), Y_e)$
        \ENDFOR
    \ENDFOR
\RETURN $\theta_\mathit{e}$ to server        
    
\end{algorithmic}
\end{algorithm}

\section{Experiments}\label{sec:exp}

\subsection{Experimental Setup}\label{ssec:exp-setup}
\subsubsection{Datasets}\label{sssec:exp-dataset}

\noindent In our experiments, we use three image datasets with different resolutions and class structures: CIFAR-100, Tiny ImageNet, and ImageNet. \textbf{CIFAR-100} contains a total of 60,000 color images, each with a resolution of 32×32 pixels. These images are uniformly distributed across 100 mutually exclusive classes, with 500 training samples and 100 test samples per class. Each image belongs to a single class, representing various objects and scenes such as "apple," "forest," and "bicycle," among others. The CIFAR-100 class hierarchy includes 22 coarse-grained classes, and the maximum depth is 5. 
\textbf{Tiny ImageNet} is a subset of ImageNet~\cite{deng2009imagenet} constructed from WordNet~\cite{miller1995wordnet} and downsized to 64×64 resolution. It consists of 200 distinct object classes, each containing 500 training images and 50 validation images, for a total of 100,000 training and 10,000 validation images. The class hierarchy of Tiny ImageNet, built using WordNet contains 37 superclasses, and the maximum depth is 6.
\textbf{ImageNet} is a large-scale benchmark of natural images organized by the WordNet hierarchy, comprising more than 14 million high-resolution images across thousands of categories; in our experiments, we used a subset of the ILSVRC 2012 training set, sampling 500,000 images via stratified sampling (500 images per class). The class hierarchy of ImageNet, built using WordNet contains 66 superclasses, and the maximum depth is 6. For the ImageNet dataset, we conducted experiments only on baselines that demonstrated notable performance on other datasets.

\subsubsection{Baselines}\label{sssec:exp-baseline}
\noindent To the best of our knowledge, prior work has not explicitly studied client-specific label granularity mismatch in federated hierarchical image classification. Therefore, we used hierarchical classification model in centralized machine learning as baselines. Their federated algorithm used FedAvg\cite{mcmahan2017communication}, the same as in FedBDFT.
The baselines are mainly divided into two types: those using only leaf labels, such as \emph{flat softmax} and parameter sharing softmax (\emph{PS softmax})~\cite{NEURIPS2022_727855c3}, and those using hierarchical labels, such as conditional sigmoid (\emph{cond sigmoid})~\cite{brust2019integrating}, conditional softmax (\emph{cond softmax})~\cite{redmon2017yolo9000}, \emph{soft-max-descendant}, and \emph{soft-max-margin}~\cite{NEURIPS2022_727855c3}.

\textbf{\emph{Flat softmax}} is a flat classifier-based hierarchical classification model that trains only on leaf classes using cross-entropy loss. 
\textbf{\emph{PS softmax}}~\cite{NEURIPS2022_727855c3} trains on leaf classes using cross-entropy loss by summing the logits of the leaf class and its ancestor classes. The prediction probability for higher-level classes is calculated as the sum of the probabilities of its child classes.
\textbf{\emph{Cond sigmoid}}~\cite{brust2019integrating} is a multi-label sigmoid classifier-based hierarchical classification model that trains each class using binary cross-entropy loss via hierarchical labels. 
\textbf{\emph{Cond softmax}}~\cite{redmon2017yolo9000} trains locally only between each pair of siblings within the class hierarchy using cross-entropy loss. The prediction probability for a class is calculated as the product of its own and all of its ancestor classes' respective local prediction probabilities.
\textbf{\emph{Soft-max-descendant}}~\cite{NEURIPS2022_727855c3} locally trains on a class and its hierarchical negative classes using cross-entropy loss for itself and its ancestor classes.
\textbf{\emph{Soft-max-margin}}~\cite{NEURIPS2022_727855c3} trains on the hierarchical negative classes of its own class but assigns a logit margin to them. The class prediction probability is calculated as the sum of the softmax values for all descendant classes over the softmax values for all classes.

\subsubsection{Evaluation Metrics}\label{sssec:exp-metric}
\noindent In flat classification, the primary evaluation criterion is \textbf{accuracy} with respect to the ground truth. Relationships among classes are ignored, and no partial credit is given unless predictions \emph{exactly} match the ground truth. By contrast, hierarchical classification accounts for relationships among classes during training and evaluation how \emph{closely} a prediction aligns with the ground truth under the given class hierarchy.

In a centralized setting, hierarchical recall and hierarchical precision were previously defined using class information content derived from class sample ratio~\cite{NEURIPS2022_727855c3} or pixel frequencies~\cite{Zhao_2017_ICCV} in the dataset. However, in label granularity skew, the information content of a class can vary depending on each client’s local label hierarchy. 
Therefore, we evaluate semantic consistency using a hierarchical F-score derived from the Wu-Palmer similarity between predicted and ground-truth labels. Following prior work~\cite{Zhao_2017_ICCV}, we compute hierarchical precision, recall, and F-score from class depths in $\mathcal{H}_{\mathrm{full}}$, the original class hierarchy guided from WordNet.

\subsubsection{Hyperparameters}\label{sssec:exp-hyperparams}
\noindent Since hierarchical classification in federated learning on general-domain image datasets has not been previously explored, we began by searching for the optimal learning rate for each baseline. We first split the original training set into training and validation subsets in an 8:2 ratio and used the original validation set as the test set. For the CIFAR-100 case, the search for baselines and FedBDFT was conducted using 5 candidate learning rates and 5 random seeds under an IID setting. Optimal learning rates were used for the label granularity skew and zero-shot learning experiments. Each experiment was conducted with another 5 to 10 random seeds, and results are reportedin the Section~\ref{ssec:exp-result} as means with 95\% confidence intervals of hierarchical F1 on the test set. Detailed hyperparameter settings for each dataset are provided in Table~\ref{tb:exp-hyperparams} in the Appendix.

\subsection{Experiment Results}\label{ssec:exp-result}

\subsubsection{Label Granularity Skewness}\label{sssec:exp-lgs}
\begin{figure*}[t]
  \centering\includegraphics[width=\linewidth]{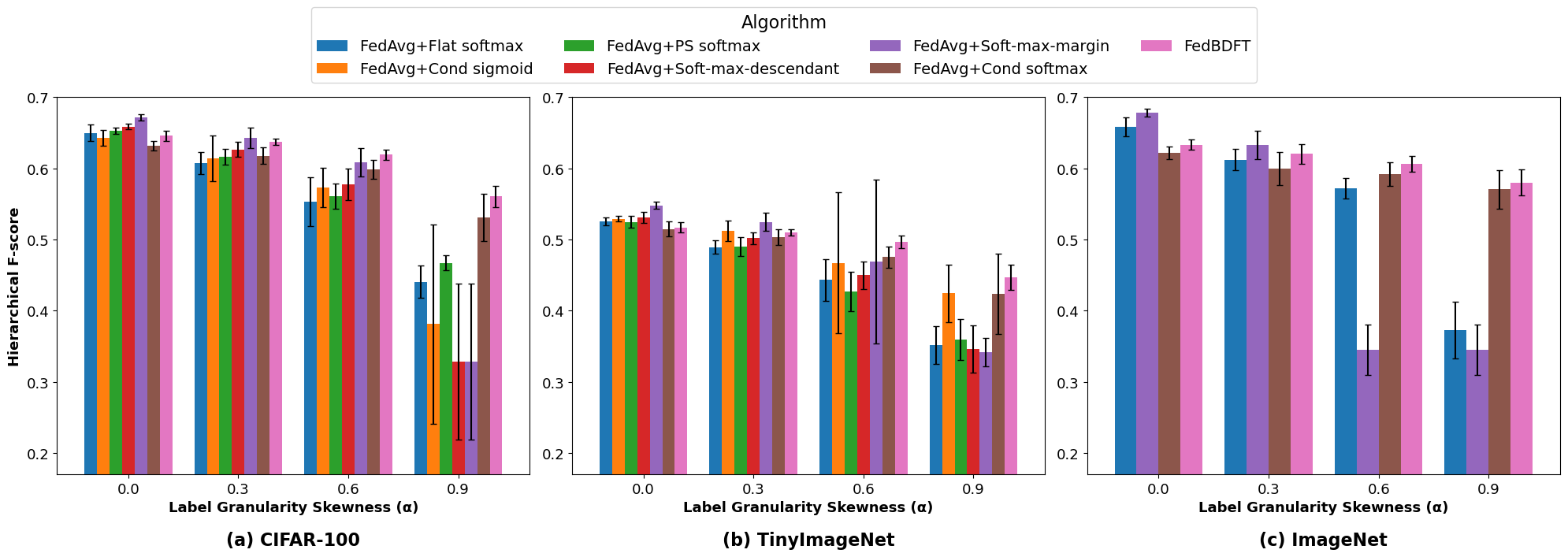}
  \caption{Performance of FedBDFT under various label granularity skewness scenarios}
  \label{fig:exp-lgs}
\end{figure*}

\noindent We first examine how the \textit{label granularity skew} adversely affects conventional federated learning and how \emph{cond softmax} and FedBDFT mitigate it. The set of label granularity skewness ($\alpha$) used in this experiment is \{0.0, 0.3, 0.6, 0.9\}. Fig.~\ref{fig:exp-lgs} illustrates an overview of the results, and Table~\ref{tab:exp-lgs-cifar100}, \ref{tab:exp-lgs-tinyimagenet}, and \ref{tab:exp-lgs-imagenet} report detailed hierarchical recall, precision, and F-score values for each dataset. In each column of the tables, bold values indicate the highest value, and underlined values indicate the second-highest value.
As skewness increases, the performance of the other baselines decreases dramatically. In contrast, although \emph{cond softmax} shows relatively lower performance in the non-skewed setting, it shows only minor performance degradation and achieves strong performance under severe skew. Furthermore, FedBDFT, an improved variant of \emph{cond softmax}, even outperforms \emph{cond softmax}.
Notably, \emph{soft-max-margin} and \emph{soft-max-descent}, which achieve the best performance at $\alpha=0.0$, fail to converge properly at $\alpha=0.9$ and occasionally also at $\alpha=0.6$. 

Fundamentally, the baselines use cross-entropy loss as their base loss function. This means that known classes are treated as both positive and negative classes, while unknown classes are trained solely as negative classes. \emph{Flat softmax} and \emph{PS softmax}, in addition to this fundamental issue, can only use flat labels, meaning fewer data points are available for training as skewness increases. \emph{Soft-max-margin} and \emph{soft-max-descendant} suffer greater adverse effects from unknown classes due to their hierarchical negative label-based loss functions, whereas \emph{cond sigmoid} faces similar issues due to its binary cross-entropy loss.
In contrast, \emph{cond softmax} is trained with a relatively decoupled loss, where the cross-entropy loss over sibling classes is computed independently for each branch. This property not only confines the influence of unknown classes to their siblings, but also better preserves the features of unknown classes obtained through aggregation, which may be known classes for other clients. This mitigates performance degradation caused by the heterogeneity.
FedBDFT achieves the highest performance at $\alpha \ge 0.6$ across all datasets by decoupling the \emph{cond softmax} feature extractor into independent local classifiers, thereby specializing each branch to its task. This design also facilitates the effective learning of generalized hierarchical feature representations. Although FedBDFT is not the top-performing model under relatively low label granularity skew (e.g., $\alpha < 0.6$), it nevertheless exhibits strong generalization ability, as evidenced by its best performance on unseen classes in zero-shot learning under the $\alpha = 0.0$ across all datasets.

\begin{table*}
\caption{Hierarchical metrics on CIFAR-100 under Label Granularity Skew}
\label{tab:exp-lgs-cifar100}
\centering
\renewcommand{\arraystretch}{1.3}
\begin{tabular}{l|ccc|ccc}
\toprule
Label Granularity Skewness ($\alpha$) & \multicolumn{3}{c}{0.0} & \multicolumn{3}{c}{0.3} \\
\midrule
Algorithm & H.Recall & H.Precision & H.F-score & H.Recall & H.Precision & H.F-score \\
\midrule
\textbf{FedAvg+Flat softmax} & 0.656 ($\pm$ 0.011) & 0.655 ($\pm$ 0.012) & 0.650 ($\pm$ 0.012) & 0.614 ($\pm$ 0.018) & 0.613 ($\pm$ 0.014) & 0.608 ($\pm$ 0.016) \\
\textbf{FedAvg+Cond sigmoid}~\cite{brust2019integrating} & 0.645 ($\pm$ 0.012) & 0.651 ($\pm$ 0.011) & 0.642 ($\pm$ 0.011) & 0.617 ($\pm$ 0.035) & 0.623 ($\pm$ 0.030) & 0.614 ($\pm$ 0.032) \\
\textbf{FedAvg+PS softmax}~\cite{NEURIPS2022_727855c3} & 0.658 ($\pm$ 0.005) & 0.659 ($\pm$ 0.005) & 0.653 ($\pm$ 0.004) & 0.622 ($\pm$ 0.014) & 0.622 ($\pm$ 0.008) & 0.616 ($\pm$ 0.011) \\
\textbf{FedAvg+Soft-max-descendant}~\cite{NEURIPS2022_727855c3} & \second{0.664 ($\pm$ 0.004)} & \second{0.664 ($\pm$ 0.004)} & \second{0.659 ($\pm$ 0.004)} & 0.633 ($\pm$ 0.011) & 0.632 ($\pm$ 0.010) & 0.627 ($\pm$ 0.010) \\
\textbf{FedAvg+Soft-max-margin}~\cite{NEURIPS2022_727855c3} & \best{0.682 ($\pm$ 0.005)} & \best{0.672 ($\pm$ 0.005)} & \best{0.672 ($\pm$ 0.005)} & \best{0.657 ($\pm$ 0.014)} & \best{0.640 ($\pm$ 0.015)} & \best{0.643 ($\pm$ 0.014)} \\
\hline
\textbf{FedAvg+Cond softmax}~\cite{redmon2017yolo9000} & 0.637 ($\pm$ 0.007) & 0.637 ($\pm$ 0.007) & 0.631 ($\pm$ 0.007) & 0.624 ($\pm$ 0.013) & 0.623 ($\pm$ 0.009) & 0.618 ($\pm$ 0.011) \\
\textbf{FedBDFT} & 0.657 ($\pm$ 0.007) & 0.645 ($\pm$ 0.007) & 0.646 ($\pm$ 0.007) & \second{0.649 ($\pm$ 0.005)} & \second{0.637 ($\pm$ 0.005)} & \second{0.637 ($\pm$ 0.004)} \\
\bottomrule
\end{tabular}
\vspace{0.5em}

\begin{tabular}{l|ccc|ccc}
\toprule
Label Granularity Skewness ($\alpha$) & \multicolumn{3}{c}{0.6} & \multicolumn{3}{c}{0.9} \\
\midrule
Algorithm & H.Recall & H.Precision & H.F-score & H.Recall & H.Precision & H.F-score \\
\midrule
\textbf{FedAvg+Flat softmax} & 0.562 ($\pm$ 0.037) & 0.557 ($\pm$ 0.033) & 0.553 ($\pm$ 0.034) & 0.447 ($\pm$ 0.029) & 0.453 ($\pm$ 0.018) & 0.441 ($\pm$ 0.023) \\
\textbf{FedAvg+Cond sigmoid}~\cite{brust2019integrating} & 0.577 ($\pm$ 0.032) & 0.582 ($\pm$ 0.027) & 0.573 ($\pm$ 0.028) & 0.375 ($\pm$ 0.150) & 0.406 ($\pm$ 0.137) & 0.381 ($\pm$ 0.140) \\
\textbf{FedAvg+PS softmax}~\cite{NEURIPS2022_727855c3} & 0.570 ($\pm$ 0.022) & 0.565 ($\pm$ 0.016) & 0.561 ($\pm$ 0.018) & 0.474 ($\pm$ 0.012) & 0.476 ($\pm$ 0.014) & 0.467 ($\pm$ 0.010) \\
\textbf{FedAvg+Soft-max-descendant}~\cite{NEURIPS2022_727855c3} & 0.585 ($\pm$ 0.024) & 0.582 ($\pm$ 0.023) & 0.577 ($\pm$ 0.022) & 0.334 ($\pm$ 0.133) & 0.350 ($\pm$ 0.079) & 0.328 ($\pm$ 0.110) \\
\textbf{FedAvg+Soft-max-margin}~\cite{NEURIPS2022_727855c3} & \second{0.624 ($\pm$ 0.015)} & \second{0.605 ($\pm$ 0.024)} & \second{0.609 ($\pm$ 0.020)} & 0.334 ($\pm$ 0.133) & 0.350 ($\pm$ 0.079) & 0.328 ($\pm$ 0.110) \\
\hline
\textbf{FedAvg+Cond softmax}~\cite{redmon2017yolo9000} & 0.609 ($\pm$ 0.015) & 0.600 ($\pm$ 0.013) & 0.599 ($\pm$ 0.013) & \second{0.545 ($\pm$ 0.034)} & \second{0.532 ($\pm$ 0.034)} & \second{0.531 ($\pm$ 0.033)} \\
\textbf{FedBDFT} & \best{0.633 ($\pm$ 0.008)} & \best{0.618 ($\pm$ 0.007)} & \best{0.619 ($\pm$ 0.007)} & \best{0.581 ($\pm$ 0.016)} & \best{0.554 ($\pm$ 0.017)} & \best{0.561 ($\pm$ 0.015)} \\
\bottomrule
\end{tabular}
\end{table*}

\begin{table*}
\caption{Hierarchical metrics on TinyImageNet under Label Granularity Skew}
\label{tab:exp-lgs-tinyimagenet}
\centering
\renewcommand{\arraystretch}{1.3}
\begin{tabular}{l|ccc|ccc}
\toprule
Label Granularity Skewness ($\alpha$) & \multicolumn{3}{c}{0.0} & \multicolumn{3}{c}{0.3} \\
\midrule
Algorithm & H.Recall & H.Precision & H.F-score & H.Recall & H.Precision & H.F-score \\
\midrule
\textbf{FedAvg+Flat softmax} & 0.532 ($\pm$ 0.005) & 0.531 ($\pm$ 0.007) & 0.525 ($\pm$ 0.006) & 0.496 ($\pm$ 0.010) & 0.496 ($\pm$ 0.012) & 0.489 ($\pm$ 0.010) \\
\textbf{FedAvg+Cond sigmoid}~\cite{brust2019integrating} & 0.510 ($\pm$ 0.004) & \best{0.577 ($\pm$ 0.006)} & 0.529 ($\pm$ 0.004) & 0.493 ($\pm$ 0.018) & \best{0.560 ($\pm$ 0.023)} & \second{0.512 ($\pm$ 0.014)} \\
\textbf{FedAvg+PS softmax}~\cite{NEURIPS2022_727855c3} & 0.532 ($\pm$ 0.008) & 0.530 ($\pm$ 0.008) & 0.525 ($\pm$ 0.008) & 0.497 ($\pm$ 0.013) & 0.496 ($\pm$ 0.014) & 0.490 ($\pm$ 0.013) \\
\textbf{FedAvg+Soft-max-descendant}~\cite{NEURIPS2022_727855c3} & \second{0.538 ($\pm$ 0.008)} & 0.536 ($\pm$ 0.009) & \second{0.531 ($\pm$ 0.008)} & 0.508 ($\pm$ 0.008) & 0.509 ($\pm$ 0.011) & 0.502 ($\pm$ 0.009) \\
\textbf{FedAvg+Soft-max-margin}~\cite{NEURIPS2022_727855c3} & \best{0.559 ($\pm$ 0.006)} & \second{0.549 ($\pm$ 0.005)} & \best{0.548 ($\pm$ 0.005)} & \best{0.536 ($\pm$ 0.011)} & \second{0.526 ($\pm$ 0.016)} & \best{0.525 ($\pm$ 0.013)} \\
\hline
\textbf{FedAvg+Cond softmax}~\cite{redmon2017yolo9000} & 0.522 ($\pm$ 0.011) & 0.520 ($\pm$ 0.010) & 0.515 ($\pm$ 0.010) & 0.513 ($\pm$ 0.010) & 0.507 ($\pm$ 0.013) & 0.503 ($\pm$ 0.011) \\
\textbf{FedBDFT} & 0.533 ($\pm$ 0.007) & 0.514 ($\pm$ 0.008) & 0.517 ($\pm$ 0.007) & \second{0.527 ($\pm$ 0.004)} & 0.506 ($\pm$ 0.005) & 0.510 ($\pm$ 0.005) \\
\bottomrule
\end{tabular}
\vspace{0.5em}
\begin{tabular}{l|ccc|ccc}
\toprule
Label Granularity Skewness ($\alpha$) & \multicolumn{3}{c}{0.6} & \multicolumn{3}{c}{0.9} \\
\midrule
Algorithm & H.Recall & H.Precision & H.F-score & H.Recall & H.Precision & H.F-score \\
\midrule
\textbf{FedAvg+Flat softmax} & 0.450 ($\pm$ 0.030) & 0.451 ($\pm$ 0.029) & 0.443 ($\pm$ 0.029) & 0.360 ($\pm$ 0.029) & 0.359 ($\pm$ 0.026) & 0.352 ($\pm$ 0.027) \\
\textbf{FedAvg+Cond sigmoid}~\cite{brust2019integrating} & 0.450 ($\pm$ 0.079) & \best{0.516 ($\pm$ 0.118)} & 0.467 ($\pm$ 0.099) & 0.358 ($\pm$ 0.062) & \best{0.557 ($\pm$ 0.026)} & \second{0.424 ($\pm$ 0.041)} \\
\textbf{FedAvg+PS softmax}~\cite{NEURIPS2022_727855c3} & 0.437 ($\pm$ 0.029) & 0.431 ($\pm$ 0.029) & 0.427 ($\pm$ 0.028) & 0.366 ($\pm$ 0.028) & 0.370 ($\pm$ 0.033) & 0.359 ($\pm$ 0.029) \\
\textbf{FedAvg+Soft-max-descendant}~\cite{NEURIPS2022_727855c3} & 0.459 ($\pm$ 0.021) & 0.455 ($\pm$ 0.020) & 0.450 ($\pm$ 0.019) & 0.357 ($\pm$ 0.033) & 0.353 ($\pm$ 0.038) & 0.346 ($\pm$ 0.033) \\
\textbf{FedAvg+Soft-max-margin}~\cite{NEURIPS2022_727855c3} & 0.480 ($\pm$ 0.122) & 0.472 ($\pm$ 0.104) & 0.469 ($\pm$ 0.115) & 0.358 ($\pm$ 0.021) & 0.342 ($\pm$ 0.032) & 0.342 ($\pm$ 0.020) \\
\hline
\textbf{FedAvg+Cond softmax}~\cite{redmon2017yolo9000} & \second{0.488 ($\pm$ 0.015)} & 0.475 ($\pm$ 0.017) & \second{0.475 ($\pm$ 0.015)} & \second{0.449 ($\pm$ 0.062)} & 0.414 ($\pm$ 0.048) & 0.423 ($\pm$ 0.056) \\
\textbf{FedBDFT} & \best{0.516 ($\pm$ 0.007)} & \second{0.492 ($\pm$ 0.012)} & \best{0.497 ($\pm$ 0.009)} & \best{0.476 ($\pm$ 0.013)} & \second{0.434 ($\pm$ 0.022)} & \best{0.447 ($\pm$ 0.018)} \\
\bottomrule
\end{tabular}
\end{table*}

\begin{table*}
\caption{Hierarchical metrics on ImageNet under Label Granularity Skew}
\label{tab:exp-lgs-imagenet}
\centering
\renewcommand{\arraystretch}{1.3}
\begin{tabular}{l|ccc|ccc}
\toprule
Label Granularity Skewness ($\alpha$) & \multicolumn{3}{c}{0.0} & \multicolumn{3}{c}{0.3} \\
\midrule
Algorithm & H.Recall & H.Precision & H.F-score & H.Recall & H.Precision & H.F-score \\
\midrule
\textbf{FedAvg+Flat softmax} & \second{0.662 ($\pm$ 0.012)} & \second{0.664 ($\pm$ 0.014)} & \second{0.658 ($\pm$ 0.013)} & 0.617 ($\pm$ 0.016) & \second{0.619 ($\pm$ 0.014)} & 0.612 ($\pm$ 0.015) \\
\textbf{FedAvg+Soft-max-margin}~\cite{NEURIPS2022_727855c3} & \best{0.687 ($\pm$ 0.006)} & \best{0.680 ($\pm$ 0.005)} & \best{0.679 ($\pm$ 0.006)} & \best{0.642 ($\pm$ 0.019)} & \best{0.634 ($\pm$ 0.020)} & \best{0.633 ($\pm$ 0.020)} \\
\hline
\textbf{FedAvg+Cond softmax}~\cite{redmon2017yolo9000} & 0.628 ($\pm$ 0.008) & 0.627 ($\pm$ 0.009) & 0.622 ($\pm$ 0.009) & 0.607 ($\pm$ 0.023) & 0.603 ($\pm$ 0.022) & 0.600 ($\pm$ 0.023) \\
\textbf{FedBDFT} & 0.645 ($\pm$ 0.006) & 0.632 ($\pm$ 0.007) & 0.633 ($\pm$ 0.007) & \second{0.633 ($\pm$ 0.012)} & 0.618 ($\pm$ 0.014) & \second{0.620 ($\pm$ 0.013)} \\
\bottomrule
\end{tabular}
\vspace{0.5em}
\begin{tabular}{l|ccc|ccc}
\toprule
Label Granularity Skewness ($\alpha$) & \multicolumn{3}{c}{0.6} & \multicolumn{3}{c}{0.9} \\
\midrule
Algorithm & H.Recall & H.Precision & H.F-score & H.Recall & H.Precision & H.F-score \\
\midrule
\textbf{FedAvg+Flat softmax} & 0.577 ($\pm$ 0.014) & 0.579 ($\pm$ 0.015) & 0.572 ($\pm$ 0.014) & 0.373 ($\pm$ 0.042) & 0.389 ($\pm$ 0.051) & 0.373 ($\pm$ 0.040) \\
\textbf{FedAvg+Soft-max-margin}~\cite{NEURIPS2022_727855c3} & 0.356 ($\pm$ 0.068) & 0.354 ($\pm$ 0.050) & 0.345 ($\pm$ 0.036) & 0.356 ($\pm$ 0.068) & 0.354 ($\pm$ 0.050) & 0.345 ($\pm$ 0.036) \\
\hline
\textbf{FedAvg+Cond softmax}~\cite{redmon2017yolo9000} & \second{0.601 ($\pm$ 0.016)} & \second{0.595 ($\pm$ 0.017)} & \second{0.592 ($\pm$ 0.017)} & \second{0.585 ($\pm$ 0.026)} & \second{0.567 ($\pm$ 0.028)} & \second{0.570 ($\pm$ 0.027)} \\
\textbf{FedBDFT} & \best{0.620 ($\pm$ 0.010)} & \best{0.604 ($\pm$ 0.012)} & \best{0.607 ($\pm$ 0.011)} & \best{0.600 ($\pm$ 0.015)} & \best{0.573 ($\pm$ 0.021)} & \best{0.580 ($\pm$ 0.018)} \\
\bottomrule
\end{tabular}
\end{table*}

\subsubsection{Zero-Shot Learning}\label{sssec:exp-zsl}
\noindent The goal of federated learning is to obtain a global model that effectively classifies all classes. 
However, in label granularity skew, each client is aware of its classes based on the local label hierarchy; accordingly, classes are divided into known classes that are learnable (seen classes) and unknown classes that are not (unseen classes). Under this setting, although the local model is trained exclusively on known classes, it should avoid distorting the features of unseen classes acquired from other clients, and thereby must learn generalized hierarchical features.
\emph{Zero-shot learning}; a task that trains only on seen classes but evaluates performance on unseen classes; can be used to analyze this capability.
FedBDFT divides a monolithic task into subtasks based on a class hierarchy, assigning specialized independent local classifiers to each. Consequently, each local classifier focuses on generalized hierarchical features within its assigned branch. 
For the experimental setup, 50\% of leaf classes are designated as seen classes, while the remaining leaf classes are designated as unseen classes and excluded from the coarsened class hierarchy $\mathcal{H}$. In other words, $\mathcal{H}$ is constructed solely from the seen classes. The seen classes are used for training and validation, and the unseen classes are used for testing. Table~\ref{tab:exp-zsl} shows the hierarchical F-scores for each algorithm in the seen and unseen classes in CIFAR-100, TinyImageNet, and ImageNet. 
Although FedBDFT does not achieve the best performance on seen classes, as in environments without label granularity skew, it attains the highest performance on unseen classes, demonstrating its ability to capture generalized hierarchical features and thereby alleviating the bias introduced by label granularity skew.

\begin{table*}
\caption{Hierarchical F-score for zero-shot learning under IID setting}
\label{tab:exp-zsl}
\centering
\renewcommand{\arraystretch}{1.3}
\begin{tabular}{l|cc|cc|cc}
\toprule
Dataset & \multicolumn{2}{c}{CIFAR-100} & \multicolumn{2}{c}{TinyImageNet} & \multicolumn{2}{c}{ImageNet} \\
Algorithm & Seen & \textbf{Unseen} & Seen & \textbf{Unseen} & Seen & \textbf{Unseen} \\
\midrule
\textbf{FedAvg+Flat softmax} & 0.681 ($\pm$ 0.018) & 0.441 ($\pm$ 0.024) & 0.540 ($\pm$ 0.016) & 0.390 ($\pm$ 0.009) & \second{0.660 ($\pm$ 0.018)} & 0.513 ($\pm$ 0.011) \\
\textbf{FedAvg+Cond sigmoid} & 0.635 ($\pm$ 0.058) & 0.412 ($\pm$ 0.040) & 0.409 ($\pm$ 0.373) & 0.332 ($\pm$ 0.275) & - & - \\
\textbf{FedAvg+PS softmax} & 0.675 ($\pm$ 0.024) & 0.445 ($\pm$ 0.017) & 0.538 ($\pm$ 0.012) & 0.396 ($\pm$ 0.013) & - & - \\
\textbf{FedAvg+Soft-max-descendant} & \second{0.696 ($\pm$ 0.014)} & 0.455 ($\pm$ 0.023) & \second{0.558 ($\pm$ 0.020)} & 0.400 ($\pm$ 0.013) & - & - \\
\textbf{FedAvg+Soft-max-margin} & \best{0.700 ($\pm$ 0.010)} & \second{0.462 ($\pm$ 0.018)} & \best{0.563 ($\pm$ 0.015)} & \second{0.410 ($\pm$ 0.015)} & \best{0.661 ($\pm$ 0.032)} & \second{0.527 ($\pm$ 0.010)} \\
\hline
\textbf{FedAvg+Cond softmax} & 0.654 ($\pm$ 0.028) & 0.442 ($\pm$ 0.022) & 0.528 ($\pm$ 0.019) & 0.395 ($\pm$ 0.011) & 0.623 ($\pm$ 0.011) & 0.512 ($\pm$ 0.015) \\
\textbf{FedBDFT} & 0.665 ($\pm$ 0.020) & \best{0.467 ($\pm$ 0.021)} & 0.531 ($\pm$ 0.019) & \best{0.415 ($\pm$ 0.019)} & 0.633 ($\pm$ 0.012) & \best{0.531 ($\pm$ 0.011)} \\
\bottomrule
\end{tabular}
\end{table*}

\section{Discussion} \label{sec:discussion}

\noindent The local label hierarchy generation algorithm is one of the key components for defining the label granularity skew setting. We employed Probabilistic Relational Neighbor to construct a moderate label hierarchy, as opposed to the naive and overly extreme alternatives. Although this node classification-based algorithm can generate a more realistic local label hierarchy than those approaches, an even more realistic hierarchy could be obtained by considering not only inter-node distances but also parent-child relations when determining the knowledge boundaries in the class hierarchy.
Moreover, this algorithm randomly assigns leaf classes as known classes. Therefore, if known classes were selected by taking into account the semantic relatedness among leaf classes, it would be possible to generate a more coherent and, consequently, more realistic local label hierarchy that better reflects the client's knowledge within the class hierarchy.

Meanwhile, WordNet-guided Hierarchy Coarsening via Silhouette Score effectively alleviates label space inflation while preserving hierarchical information by identifying the optimal children set for each branch. However, since this procedure requires comparing the clustering quality of all possible combinations of child classes, generating a coarsened label hierarchy becomes increasingly time-consuming as the number of leaf classes grows, even when a bottom-up strategy is adopted to reduce the search space.

\section{Conclusion} \label{sec:conclusion}

\noindent This study departs from the conventional assumption that clients, as data owners, possess complete knowledge when annotating data for hierarchical classification. Instead, we consider a more realistic setting in which clients have incomplete and heterogeneous knowledge. Under such conditions, the granularity of hierarchical labeling assigned to each class may vary across clients, and we are the first to identify it as label granularity skew.
To model heterogeneous client knowledge, we propose a local label hierarchy construction algorithm based on Probabilistic Relational Neighbor. The proposed algorithm is designed to capture balanced decision boundaries between two extremes, namely overly contracted and overly expanded knowledge boundaries.

In addition, WordNet-guided Hierarchy Coarsening via Silhouette Score constructs a class hierarchy that more faithfully reflects hierarchical inter-class relationships while mitigating excessive expansion of the label space. Unlike prior studies that relied on hierarchies derived from limited domains or inaccurately manipulated class structures for experimental convenience, our approach yields a more realistic and semantically coherent hierarchy.

In this heterogeneity, we experimentally show that models that strongly enforce hierarchical feature learning across classes often fail to converge. In contrast, \emph{cond softmax}, which learns hierarchical characteristics more loosely, proves effective in the heterogeneous environment. Furthermore, we demonstrate that FedBDFT, which leverages this property, is particularly effective under highly heterogeneous conditions.

\bibliographystyle{ieeetr}
\bibliography{refs}

\clearpage
\section{Appendix}

\subsection{Experiment Environments}
\noindent We utilized two different environments for the small-scale dataset (CIFAR-100 and TinyImageNet) and the large-scale dataset (ImageNet). 

\textbf{Hyperparameters and System Details}: Table~\ref{tb:exp-hyperparams} shows the details of the hyperparameters for both the small-scale data set and the large-scale data set. In large-scale dataset experiments, batch size is increased, and client joint ratio, number of experiments for searching for the optimal learning rate and training are reduced. We additionally use a random seed to sample a subset of the training set for searching for the optimal learning rate and training individually. 
Table~\ref{tb:exp-system} shows the hardware and system versions used for the experiments.

\begin{table}[htbp]
\centering
\caption{Hyperparameters for Experiments}
\label{tb:exp-hyperparams}
\renewcommand{\arraystretch}{1.5}
\begin{tabular}{|l|c|}
\hline
\multicolumn{2}{|c|}{\textbf{Shared Hyperparameters}} \\
\hline
The number of clients & 10 \\
\hline
The number of local epoch & 1 \\
\hline
Early stopping patience & 10 \\
\hline
Early stopping minimum delta & 1e-4 \\
\hline
Label granularity skewness ($\alpha$) & \{0.0, 0.3, 0.6, 0.9\} \\
\hline
Unseen class ratio for zero-shot learning & 0.5 \\
\hline

\hline
\multicolumn{2}{|c|}{\textbf{CIFAR-100 \& TinyImageNet}} \\
\hline
Client joint ratio & 0.5 \\
\hline
Batch size            & 32 \\
\hline
Random seed for searching for optimal learning rate & [42 - 46] \\
\hline
Random seed for training & [47 - 56] \\
\hline

\hline
\multicolumn{2}{|c|}{\textbf{ImageNet}} \\
\hline
Client joint ratio & 0.4 \\
\hline
Batch size            & 128 \\
\hline
Random seed for sampling a subset of training set & 42, 46 \\
\hline
Random seed for searching for optimal learning rate & [43 - 45] \\
\hline
Random seed for training & [47 - 51] \\
\hline
\end{tabular}
\end{table}

\begin{table}[htbp]
\centering
\caption{System Details for Experiments}
\label{tb:exp-system}
\renewcommand{\arraystretch}{1.5}
\begin{tabular}{|l|c|}
\hline
\multicolumn{2}{|c|}{\textbf{CIFAR-100 \& TinyImageNet}} \\
\hline
Processor & Intel(R) Core(TM) i9-10980XE CPU @ 3.00GHz \\
\hline
GPU & 2 NVIDIA GeForce RTX 3090 \\
\hline
System version & Ubuntu 22.04.5 LTS and CUDA 12.6 \\
\hline

\hline
\multicolumn{2}{|c|}{\textbf{ImageNet}} \\
\hline
Processor & Intel(R) Xeon(R) Gold 5218R CPU @ 2.10GHz \\
\hline
GPU & 2 NVIDIA GeForce RTX 4090 \\
\hline
System version & Ubuntu 24.04.3 LTS and CUDA 13.0\\
\hline
\end{tabular}
\end{table}

\textbf{Base Models}:
We adopted two modified architectures: the CNN model used in FedAvg~\cite{mcmahan2017communication} for small-scale datasets and SqueezeNet~\cite{iandola2016squeezenetalexnetlevelaccuracy50x} for large-scale datasets. Tables~\ref{tb:model_receipt_small} and~\ref{tb:model_receipt_large} provide detailed base model configurations for each dataset. In hierarchical classification, the dimensionality of the output logits may vary across hierarchical models because each employs a distinct hierarchical output activation function. For example, flat softmax computes the probability of a class by summing the probabilities of its leaf descendants; therefore, the base model must output logits for all leaf classes. In contrast, each base model (local classifier) in BDFT requires only the logits of the child classes associated with its class, reflecting its divide-and-conquer approach. Thus, the dimensionality of the output channels should be aligned with the architecture of the hierarchical model; in the tables, it is currently specified as the number of leaf classes for each dataset.

\begin{table}[t]
\centering
\caption{Base Model Configurations for Small-Scale Datasets}
\label{tb:model_receipt_small}
\begin{tabular}{llccc}
\toprule
\multicolumn{5}{c}{\textbf{CIFAR-100}} \\
\toprule
Stage   & Layer     & Kernel    & Channel   & Stride \\
\midrule
conv1   & Conv2d    & 5x5       & 32        & 1 \\
        & ReLU      & -         & -         & - \\
        & MaxPool2d & 2x2       & -         & 2 \\
\midrule
conv2   & Conv2d    & 5x5       & 64        & 1 \\
        & ReLU      & -         & -         & - \\
        & MaxPool2d & 2x2       & -         & 2 \\
\midrule
classifier & Flatten   & -         & -         & - \\ 
        & Linear    & -         & 100       & - \\
\bottomrule
\toprule
\multicolumn{5}{c}{\textbf{TinyImageNet}} \\
\toprule
Stage   & Layer     & Kernel    & Channel   & Stride \\
\midrule
conv1   & Conv2d    & 5x5       & 32        & 1 \\
        & ReLU      & -         & -         & - \\
        & MaxPool2d & 2x2       & -         & 2 \\
\midrule
conv2   & Conv2d    & 5x5       & 64        & 1 \\
        & ReLU      & -         & -         & - \\
        & MaxPool2d & 2x2       & -         & 2 \\
\midrule
conv3   & Conv2d    & 5x5       & 128        & 1 \\
        & ReLU      & -         & -         & - \\
        & MaxPool2d & 2x2       & -         & 2 \\
\midrule
classifier & Flatten   & -         & -         & - \\ 
        & Linear    & -         & 200       & - \\
\bottomrule
\end{tabular}
\end{table}

\begin{table}[t]
\centering
\caption{Base Model Configurations for Large-Scale Datasets}
\label{tb:model_receipt_large}
\begin{tabular}{llccc}
\toprule
\multicolumn{5}{c}{\textbf{ImageNet}} \\
\toprule
Stage   & Layer     & Kernel     & Channel   & Stride \\
        &           & $s_{1x1}, e_{1x1}, e_{3x3}$ &  & \\
\midrule
conv1   & Conv2d    & 3x3           & 64        & 2 \\
        & ReLU      & -             & -         & - \\
        & MaxPool2d & 3x3           & -         & 2 \\
\midrule
fire1   & Fire      & 16, 64, 64    & 64        & - \\
        & MaxPool2d & 3x3           & -         & 2 \\
\midrule
fire2   & Fire      & 32, 128, 128  & 128       & - \\
        & MaxPool2d & 3x3           & -         & 2 \\
\midrule
fire3   & Fire      & 48, 192, 192  & 256        & - \\
\midrule
classifier & Dropout   & 0.5       & -         & 1 \\
        & Conv2d    & 5x5        & 1000      & 1 \\
        & ReLU      & -          & -         & - \\
        & AdaptiveAvgPool2d      & 1x1   & - & - \\
        & Flatten   & -         & -         & - \\ 
\bottomrule
\end{tabular}
\end{table}

\end{document}